\documentclass[10pt,journal,compsoc]{IEEEtran}

\usepackage{fancybox}
\usepackage{enumitem}
\usepackage{soul}
\usepackage{color}
\usepackage{url}
\usepackage[running, mathlines]{lineno}
\usepackage{braket}
\usepackage{amsfonts}
\usepackage{amsmath}
\usepackage{graphicx}
\usepackage{caption}
\usepackage{subcaption}
\newcommand\Tstruts{\rule{0pt}{2.6ex}}
\newcommand\Bstruts{\rule[-1.3ex]{0pt}{0pt}}
\usepackage{wrapfig}
\usepackage{stfloats}

\begin{document}

\title{Domain Elastic Transform: Bayesian Function Registration for High-Dimensional Scientific Data}

\author{Osamu~Hirose and Emanuele~Rodol\`a 
  \thanks{O. Hirose is with the Institute of Science and Engineering,
          Kanazawa University, Kakuma, Kanazawa, Ishikawa 920-1192, Japan. (Corresponding author: hirose@se.kanazawa-u.ac.jp)}
  \thanks{E. Rodol\`{a} is with the Department of Computer Science, Sapienza University of Rome, Italy.}
}



\maketitle

\begin{abstract}
  Nonrigid registration is conventionally divided into point set registration, 
which aligns sparse geometries, and image registration, which aligns continuous 
intensity fields on regular grids. However, this dichotomy creates a bottleneck
for emerging scientific data, such as spatial transcriptomics, where high-dimensional
vector-valued functions, e.g., gene expression, are defined on irregular,
sparse manifolds. Consequently, researchers currently face a forced choice:
either sacrifice single-cell resolution via voxelization to utilize image-based tools,
or ignore the functional signal to utilize geometric tools.
 
To resolve this dilemma, we propose Domain Elastic Transform (DET), a grid-free
probabilistic framework that combines geometric and functional alignment. 
By treating data as functions on irregular domains, DET registers high-dimensional 
signals directly without binning. We formulate the problem within a generalized
Bayesian framework, modeling domain deformation as an elastic motion guided 
by a joint spatial-functional likelihood. 
The method is fully unsupervised and
scalable through registration on sampled points followed by displacement interpolation.
  
We evaluate DET on spatial-transcriptomics registration tasks using MERFISH
mouse-brain slices and Stereo-seq mouse-embryo atlases. 
On a 90-case MERFISH benchmark under severe perturbations without prior initialization,
DET achieved the strongest spatial overlap and topology among the evaluated pipelines,
while an accelerated PASTE2 variant achieved the highest label-transfer ARI.
In an atlas-scale MOSTA feasibility study without cross-stage ground truth,
nonrigid refinement improved several within-pipeline anatomical-domain and
boundary-consistency measures.
The results suggest that grid-free function registration is a useful complement
to existing point-set-based, image-based, and optimal-transport approaches for high-dimensional
scientific data. The implementation of DET is available at {https://github.com/ohirose/bcpd}.

\end{abstract}


\section{Introduction}
\IEEEPARstart{N}{onrigid} registration---the process of estimating a transformation that aligns two
datasets into a common coordinate system---is a cornerstone of pattern analysis and
computer vision. The breadth of this field is documented in extensive surveys covering
shape correspondence \cite{Sahillioglu20}, point cloud registration \cite{Tam13},
local surface features \cite{Guo16}, and large-scale terrestrial scanning \cite{Dong20}.
While the applications are diverse, ranging from reconstructing dynamic 3D scenes to
tracking tissue deformation, the fundamental objective remains the same: to establish
semantic correspondence between distinct structures.

\subsection{The Dichotomy of Registration Methods}
Historically, this problem has been addressed through two distinct paradigms:
\textit{point set registration} and \textit{image registration}. The former class of methods,
such as coherent point drift \cite{Myronenko10} and its Bayesian generalizations
\cite{Hirose21,Hirose21a,Hirose23}, excels at aligning geometric structures by treating
data as sparse point clouds. The versatility of these frameworks is
evidenced by their success in disparate domains, from surgical robotic blood removal
\cite{Su21} and biological landmarking \cite{Porto21} to reconstructing
human ear shapes \cite{Valdeira21}.
Despite their geometric robustness \cite{Min21, Zhang22}, these methods rely
solely on spatial coordinates, ignoring \textit{functional} signals such as color,
texture, or biological properties carried by the points.

Conversely, image registration methods exploit these continuous intensity fields to
drive alignment, often utilizing diffeomorphic flows \cite{Vercauteren09,
Ashburner07, Beg05} or optical flow \cite{Horn81}. Yet, these methods fundamentally
assume the data exist on a regular Euclidean grid, i.e., pixels or voxels. While
\textit{functional map} frameworks \cite{Ovsjanikov12} attempt to bridge this gap by
aligning spectral signatures \cite{Sun09}, 
they typically rely on spectral representations and additional optimization
steps for recovering dense or partial pointwise correspondences~\cite{Vestner17,Rodola17},
and they often assume manifold meshes that are unavailable for raw scientific
point clouds.

\subsection{The Challenge of Emerging Scientific Data}
The dichotomy between ``geometry-only'' and ``grid-only'' methods has become a
practical limitation in modern science. Emerging technologies in \textit{spatial
transcriptomics}, e.g., Stereo-seq \cite{Chen22} and Slide-seq \cite{Rodriques19},
generate datasets that are neither simple shapes nor standard images. Instead,
they represent high-dimensional vector-valued functions, i.e., gene expression,
defined on irregular, sparse point sets. 
For instance, Multiplexed Error-Robust Fluorescence in situ Hybridization
(MERFISH) can measure, at cellular resolution, high-plex gene-expression panels
ranging from hundreds to thousands of genes depending on the experiment~\cite{Zhang23}.
Applying standard image registration to such data necessitates \textit{binning} or
rasterization---aggregating discrete points into a regular grid~\cite{Clifton23STalign}.
This process is inherently lossy: it destroys cellular resolution, introduces
quantization artifacts, and obscures fine-grained anatomical boundaries.
On the other hand, applying standard point set registration fails because the geometry alone
is often ambiguous; distinct biological tissues may share similar shapes, and
correspondence can only be disambiguated by the high-dimensional functional signal
\cite{Zeira22,PASTE2GenomeRes23}.

\subsection{The Training-Free Imperative}
To address such complex tasks, the field has largely shifted toward deep learning.
Geometric transformers~\cite{Qin23,Yu23RoITr}, implicit neural representations \cite{Xie21NeuralFields},
learning-based image-registration models such as VoxelMorph~\cite{Balakrishnan19},
and trainable point-cloud registration models such as PointNetLK~\cite{Aoki19} have
achieved strong performance on benchmark datasets. 
However, these paradigms are difficult to deploy in some scientific-discovery settings.
Learning-based registration methods typically require a sufficiently large training
collection from a related distribution, even when they require no manual correspondence annotations.
 
This dependency creates a ``zero-shot gap'' in domains such as embryology or paleontology,
where each specimen may be unique or rare, making the curation of training sets impractical.
Furthermore, deep models may suffer from domain shift when deployed outside their training distribution
\cite{Yoon23DomainGen}, a concern that becomes especially relevant in rare-specimen settings.
Consequently, there remains a need for \textit{training-free} algorithms capable
of registering complex functions immediately, effectively bypassing the data-curation
bottleneck associated with deep learning.

\subsection{Contributions}
In this study, we propose \textit{Domain Elastic Transform} (DET), a generalized Bayesian
algorithm that registers vector-valued functions directly on sparse point sets.
We model domain deformation as an elastic motion and solve the inverse problem
via variational inference.
Unlike image registration methods that require lossy binning, DET utilizes
both point locations and high-dimensional functional signals without binning,
preserving cellular resolution.
The main contributions of this paper are summarized as follows:

\begin{itemize}
    \item \textbf{Grid-Free Function Registration:} We formulate registration of
    high-dimensional vector-valued functions directly on irregular point sets,
    keeping deformation in the spatial domain while functional signals guide
    correspondence.

    \item \textbf{Training-Free and Unsupervised:} Our method requires no training
    data. We demonstrate its capability to register complex biological forms in
    ``$N=1$'' regimes where annotated datasets are unavailable.

    \item \textbf{Joint Spatial-Functional Likelihood:} We introduce a
    probabilistic likelihood that couples geometric proximity with functional
    similarity without warping the functional coordinates themselves.

    \item \textbf{Dimension-Balanced Hierarchical Inference:} A dimensionality
    weight prevents high-dimensional functional signals from overwhelming spatial
    information, while functional annealing supports global pose recovery followed
    by coherent nonrigid refinement.

    \item \textbf{Scalable Implementation and Evaluation:} The acceleration
    framework inherited from BCPD~\cite{Hirose21,Hirose21a,Hirose23} makes DET
    practical for atlas-scale point sets. We evaluate the resulting method on a
    90-case MERFISH benchmark and an atlas-scale MOSTA feasibility study.
\end{itemize}


The remainder of this paper is organized as follows.
Section 2 reviews related work.
Section 3 introduces DET, and
Section 4 presents additional techniques.
Section 5 reports numerical experiments,
Section 6 discusses the results, and
Section 7 concludes the paper.


\section{Related Work}
\label{sec:related_work}

This section positions DET among spatial-transcriptomics baselines and broader methods.

\subsection{Feature-Aware Point-Cloud Registration}
Feature-aware point-cloud registration improves correspondence estimation using local
geometric descriptors, learned features, or attention-based matching. Representative methods
include FCGF~\cite{Choy19}, Deep Global Registration~\cite{Choy20}, and
PREDATOR~\cite{Huang21}, which learn geometric features, correspondence confidences,
or overlap-aware points for rigid registration of 3D scans. These methods demonstrate
the value of learned features for registration, especially in noisy or low-overlap scan pairs,
but they are primarily designed for rigid 3D scenes and rely on training data for feature or
correspondence learning. 
Since DET is fully unsupervised and training-free, it is suitable for ``$N=1$'' scientific
settings where no related training collection is available.

\subsection{Functional Maps and Shape Correspondence}
Functional-map methods also use signals on domains to guide correspondence
estimation. They represent maps in a reduced spectral basis and have been
extended with deep structured prediction, learned geometric features, and
regularized map extraction for dense shape correspondence~\cite{Litany17FMNet,Donati20DeepShells}.
In contrast, DET estimates correspondences directly from observed point coordinates and measured
feature vectors.  This explicit use of point proximity and local deformation coherence is important
when estimating spatially local deformations, such as those caused by tissue growth or local cellular
rearrangement in spatial transcriptomics data.

\subsection{Spatial-Transcriptomics Registration}
PASTE~\cite{Zeira22} and PASTE2~\cite{PASTE2GenomeRes23} are direct optimal-transport baselines
for spatial-transcriptomics slice alignment.
PASTE aligns slices by coupling spatial distance with gene-expression similarity, whereas
PASTE2 extends this framework to partial alignment, allowing partially overlapping slices
and slice-specific cell types.
Their transport plans provide barycentric
mappings and can provide strong semantic matching, but they impose no smoothness prior on the
source domain deformation. This distinction is important when the registration
output is expected to preserve local tissue neighborhoods rather than provide
high-quality semantic correspondence.
STalign \cite{Clifton23STalign} represents a complementary image-based paradigm. It rasterizes spatial
molecular data and applies large-deformation diffeomorphic metric mapping to the resulting images.
It is therefore a natural diffeomorphic baseline for spatial transcriptomics, while also testing
the effect of the grid-conversion step that DET avoids.

\subsection{Spatial-Omics Integration}
Several spatial-omics methods map or integrate cells and expression profiles without necessarily
estimating a pairwise deformation between two observed spatial domains. novoSpaRc~\cite{Nitzan19NovoSpaRc}
reconstructs spatial organization from single-cell expression profiles,
SpaOTsc~\cite{Cang20SpaOTsc} infers spatial and signaling relationships between cells,
and Tangram~\cite{Biancalani21Tangram} maps single-cell transcriptomes to spatial data.

GPSA~\cite{Jones23GPSA} is more directly related to spatial registration. It aligns
spatial genomics samples by warping them into a common coordinate system and modeling
phenotypic readouts there with a deep Gaussian-process model. In contrast, DET performs
pairwise function registration, estimating a coherent source-to-target deformation with
explicit spatial-functional correspondence probabilities. Computationally, GPSA relies
on sparse inducing-point inference, whereas DET controls the dominant registration cost
by downsampling and displacement field interpolation.

Developmental spatial omics introduces further complications because tissue mass and cell
populations can vary across stages. DeST-OT~\cite{Halmos25DeOTST} addresses this setting using optimal
transport with explicit growth and death effects. This objective is complementary to our
Stereo-seq demonstration: DeST-OT models distributional change across developmental stages,
whereas DET estimates a smooth deformation of a source domain under a motion coherence prior.

\section{Methods} \label{sc:bcpd}
This section describes how we derive DET.

\subsection{Problem Definition} \label{sec:meth:prob}
\begin{figure}[h]
  \vspace{-1mm}
  \centering
  \includegraphics[width=.42\textwidth]{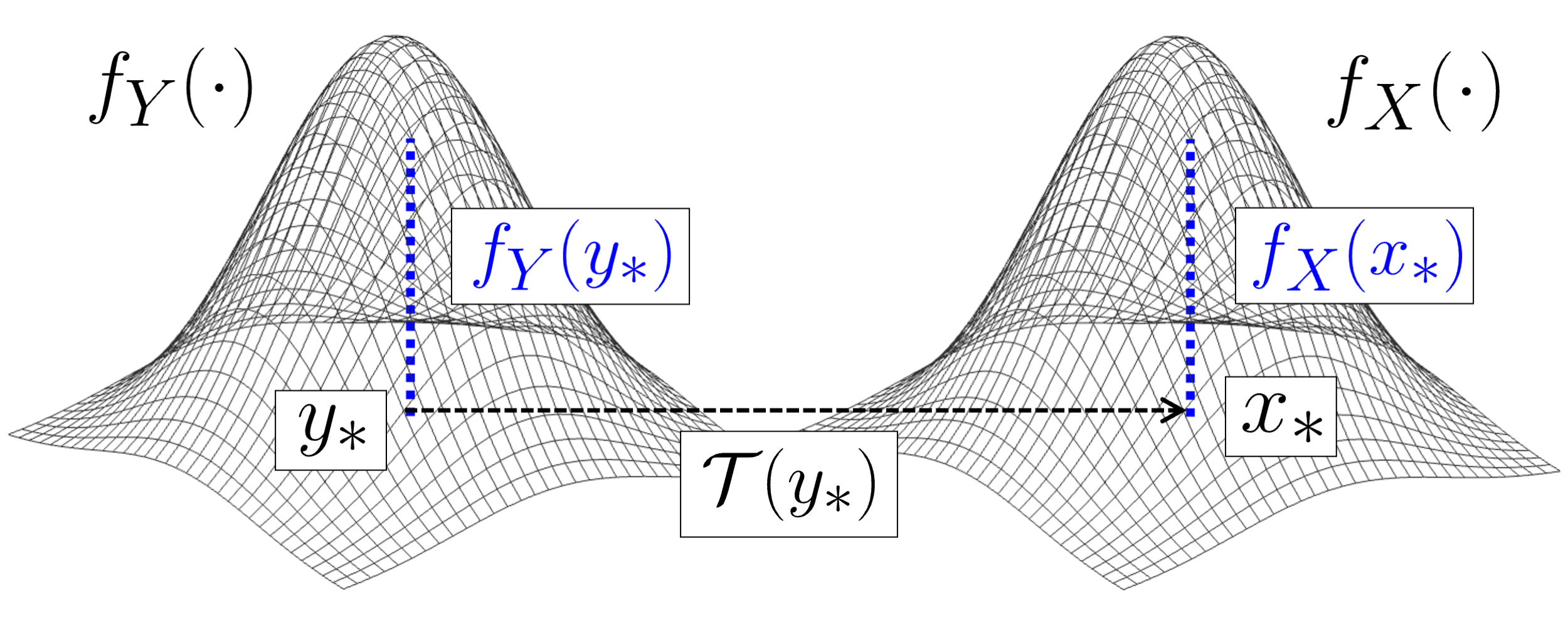}
  \vspace{-2mm}
  \caption{
    {\bf Illustration of function registration} with $D=2$ and $D'=1$.
    We find a map $\mathcal{T}$ and corresponding points $\{(x_*,y_*)\}$
    satisfying $x_*=\mathcal{T}(y_*)$ and $f_X(x_*) = f_Y(y_*)$.
  } \label{fig:prob-def}
\end{figure}

We define function registration as the process of overlaying two functions by
continuously deforming a function's domain, which may be a manifold, such as
a shape surface. Given the following multivariate and
vector-valued functions:
\begin{equation}
    f_X(\cdot) \colon \mathbb{R}^D \to \mathbb{R}^{D'}, \quad
    f_Y(\cdot) \colon \mathbb{R}^D \to \mathbb{R}^{D'}, \nonumber
\end{equation}
we seek a map $\mathcal{T}$ satisfying the following equations:
\begin{equation}
    f_X(x_*) = f_Y(y_*), \quad x_* = \mathcal{T}(y_*), \nonumber
\end{equation}
where $x_*$ and $y_*$ are corresponding points associated by the map
$\mathcal{T}:\mathbb{R}^D\rightarrow\mathbb{R}^D$, which is typically nonlinear.
Hereafter, we refer to $f_X(\cdot)$ and $f_Y(\cdot)$ as {\em target} and {\em source}
functions, respectively. Fig.~\ref{fig:prob-def} illustrates the definition of
function registration. 
\vspace{-3mm}

\subsection{Importance of Spatial-Functional Decoupling}
The above formulation differs fundamentally from applying standard point set registration
to augmented vectors, i.e., $y'_m = (y_m^T, f_Y(y_m)^T)^T$ and $x'_n = (x_n^T, f_X(x_n)^T)^T$. 
Such a naive approach would estimate a transformation in the joint space 
$\mathbb{R}^{D+D'}$, implicitly allowing the ``warping'' of the signal space 
itself, e.g., rotating gene expression profiles into spatial coordinates.
In contrast, this problem setting rigorously decouples domain deformation from signal 
identity: we seek a transformation $\mathcal{T}$ that operates exclusively on 
the spatial domain $\mathbb{R}^D$, constrained by the matching of signals in 
$\mathbb{R}^{D'}$ without deforming them.

\subsection{Outline of Our Approach} \label{sec:meth:outl}

We formulate function registration as a constrained nonrigid point set registration problem;
we use function values as constraints to address the limitations of purely geometric methods.

\subsubsection{Discretized Function Registration}
We discretize input functions to formulate function registration as point set
registration. Let us denote the discretized domains of $f_X$ and $f_Y$ by
$\{x_n\}_{n=1}^N$ and $\{y_m\}_{m=1}^M$, respectively. Then, we find a map
$\mathcal{T}$ and corresponding points $\{(x_n,y_m)\}$ that satisfy the following
conditions:
\begin{align}
  f_X(x_n)  &\approx f_Y(y_m),  ~~~x_n \approx \mathcal{T}(y_m). \nonumber
\end{align}
Here, we interpret $x_n\approx \mathcal{T}(y_m)$ as point set registration and
$f_X(x_n)\approx f_Y(y_m)$ as its constraint. Unlike the definition in
Section \ref{sec:meth:prob}, we replace the equalities with approximations. This is
because there may be no pairs of points satisfying $x_n = \mathcal{T}(y_m)$ or
$f_X(x_n) = f_Y(y_m)$ due to the discretization or the inherent differences between
functions.

\subsubsection{Probabilistic Approach}
We formulate function registration in a Bayesian setting.
Suppose a set of unknown
variables $\theta$ encodes the map $\mathcal{T}$ and corresponding points
$\{(x_n,y_m)\}$. We estimate $\theta$ on the basis of the maximum a posteriori
principle:
\begin{align}
  \hat{\theta} = \text{argmax}_{\theta} ~p(\theta|x,y,F_x,F_y), \nonumber
\end{align}
where the observations $(x,F_x)$ and $(y,F_y)$ are the discretized versions of $f_X$
and $f_Y$, defined in Table \ref{tab:notation}. Owing to the generalized Bayesian setting, we can
efficiently compute an approximate estimate of $\theta$ using an inference technique called
variational inference \cite{Bishop06}.
The next section gives a generative interpretation of the model
components and defines the corresponding generalized probabilistic
model for inference.

\subsection{Generative Model Definition} \label{sec:meth:gmdl}

\begin{table*}
\caption{Notation Used for the Generative Model.} \vspace{0mm}
\begin{tabular}{ll}  \hline
  Symbol(s)  & Definition/Description \\ \hline
             & \vspace{-2.5mm}\\ 
  $N, ~M$    & Numbers of points in the discretized domains of $f_X$ and $f_Y$, respectively. \vspace{.2mm}\\
  $D, ~D'$   & Dimensions corresponding to the function domain and codomain, i.e.,
               $f_X:\mathbb{R}^D\rightarrow\mathbb{R}^{D'}$ and $f_Y:\mathbb{R}^D\rightarrow\mathbb{R}^{D'}$. \vspace{.5mm}\\
  $I_L, ~1_L$& The unit matrix of size $L$ and the vector of all $1$s of size $L$. The size $L$ can be $D$, $D'$, $M$, or $N$.\vspace{.5mm} \\
  $x_n$      & $x_n=(x_{n1},\cdots,x_{nD})^T\in\mathbb{R}^{D}$. The $n$th point in the discretized target domain $\{x_1,\cdots,x_N\}$. \vspace{.5mm}\\
  $y_m$      & $y_m=(y_{m1},\cdots,y_{mD})^T\in\mathbb{R}^{D}$. The $m$th point in the discretized source domain $\{y_1,\cdots,y_M\}$. \vspace{.5mm}\\
  $x$        & $x=(x_1^T,\cdots,x_N^T)^T\in\mathbb{R}^{DN}$. Vector representation of a discretized target domain $\{x_1,\cdots,x_N\}$. \vspace{.5mm} \\
  $y$        & $y=(y_1^T,\cdots,y_M^T)^T\in\mathbb{R}^{DM}$. Vector representation of a discretized source domain $\{y_1,\cdots,y_M\}$. \vspace{.5mm}\\
  $F_x$      & $F_x=(f_X(x_1),\cdots,f_X(x_N))\in\mathbb{R}^{D'\times N}$. Matrix collecting the target function values. \vspace{.5mm}\\
  $F_y$      & $F_y=(f_Y(y_1),\cdots,f_Y(y_M))\in\mathbb{R}^{D'\times M}$. Matrix collecting the source function values. \vspace{.5mm}\\
  $v$        & $v=(v_1^T,\cdots,v_M^T)^T\in\mathbb{R}^{DM}$. Nonlinear displacements discretizing $v_Y(\cdot)$, 
               where $v_m\in\mathbb{R}^D$ is the displacement of $y_m$. \vspace{.5mm}\\
  $c$        & $c=(c_1,\cdots,c_N)\in\{0,1\}^N$. Outlier indicators;
               $c_n=0$ specifies $x_n$ is an outlier, and $c_n=1$ specifies $x_n$ is a non-outlier. \vspace{.5mm}\\
  $e$        & $e=(e_1,\cdots,e_N)\in \{1,\cdots,M\}^{N}$. Index variables; $e_n=m$ indicates that $x_n$ corresponds to $y_m$; $e_n$ is ignored when $c_n=0$ \vspace{.5mm}\\
  $\alpha$   & $\alpha=(\alpha_1, \cdots,\alpha_M)\in [0,1]^{M}$. Mixing probabilities;
               $\alpha_m$ is the probability that $e_n=m$, satisfying $\sum_{m=1}^M \alpha_m=1$. \vspace{.5mm}\\
  $\xi$      & $\xi=(s,R,t)$, which defines the similarity transformation $T(z)=sRz+t$ for a vector $z\in\mathbb{R}^D$. \vspace{.5mm}\\
  $\sigma^2$ & Variance $(>0)$, interpreted as the strength of the positional constraint $x_n\approx \mathcal{T}(y_m)$. \vspace{.5mm}\\
  $\Pi$      & Covariance matrix of size $D'\times D'$, which controls the strength of the function value constraint $f_X(x_n)\approx f_Y(y_m)$. \vspace{.5mm}\\
  $\theta$   & $\theta=(v,\alpha,c,e,\xi,\sigma^2,\Pi)$. Random variables mediating the target function generation. \vspace{.5mm}\\
  $\phi(z;\mu,S)$ & Gaussian distribution of $z$ with a mean $\mu$ and a covariance $S$, i.e.,
             $\phi(z;\mu,S)=|2\pi S|^{-1/2}\exp\{-\frac{1}{2} (z-\mu)^TS^{-1}(z-\mu)\}$.  \vspace{.5mm}\\ \hline
\end{tabular} \label{tab:notation}
\vspace{0mm}
\end{table*}

This section defines the model $p(x,y,F_x,F_y,\theta)$,
which provides a generative interpretation of $(x,F_x)$
given $(y,F_y)$. Table~\ref{tab:notation} defines the symbols used hereafter.

\subsubsection{Target Function Generation}
We assume a discretized source function $(y,F_y)$ generates a discretized target
function $(x,F_x)$ as follows:

\vspace{1mm}\noindent {\bf Assumptions}.
\begin{enumerate}
\item  A map $\mathcal{T}$ is generated from a motion coherence prior.
\item  A variable $c_n$ takes the value either $0$ or $1$, indicating an outlier
       or a non-outlier, with the probabilities $\omega$ and $1-\omega$, respectively.
       An outlier is a target-domain point with no corresponding source-domain point.
\item  If $c_n=0$, $(x_n, f_X(x_n))$ is generated from an outlier distribution
       $p_{\text{out}}(\cdot)$.
\item  If $c_n=1$, a variable $e_n\in\{1,\cdots,M\}$ takes the value $m$ with
       probability $\alpha_m$, satisfying $\sum_{m=1}^M \alpha_m=1$.
       The variable $e_n=m$ indicates $x_n$ corresponds to $y_m$.
\item  The location $x_n$ is generated from a $D$-dimensional normal distribution
       $\phi(x_n;\mathcal{T}(y_m),\sigma^2I_D)$.
\item  Also, $f_X(x_n)$ is generated from a $D'$-dimensional normal distribution
       $\phi(f_X(x_n);f_Y(y_m),\Pi)$. \label{item:constraint}
\item  The discretized target function $\{(x_n,f_X(x_n))\}_{n=1}^N$ is
       generated by repeating steps 2--6 $N$ times.
\end{enumerate}

\begin{figure}
  \centering
  \includegraphics[width=0.40\textwidth]{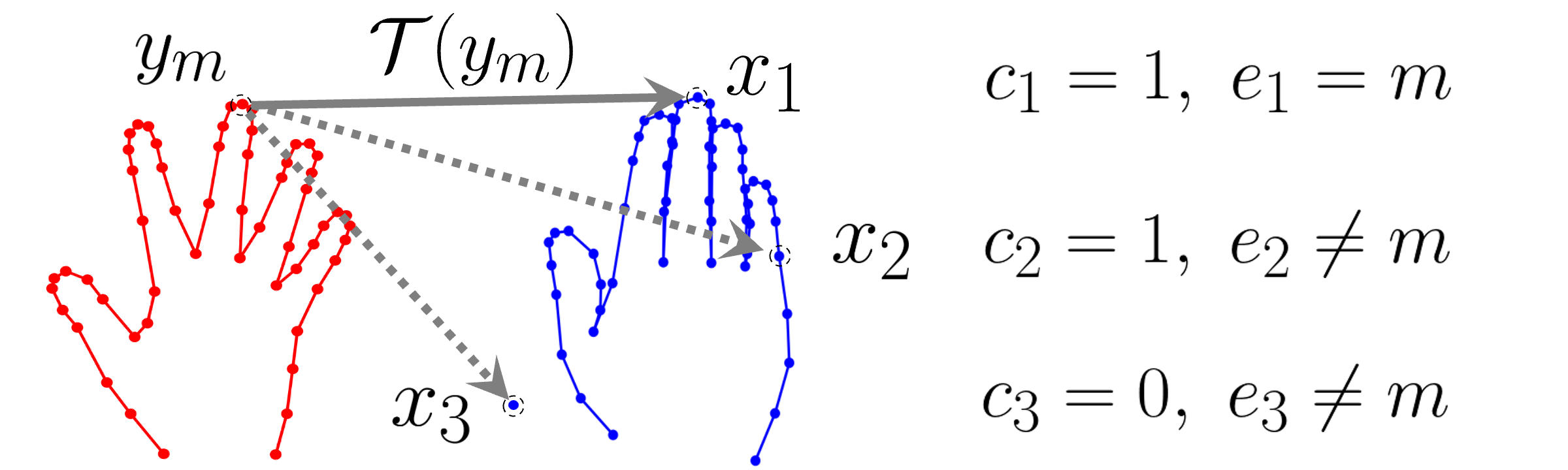}
  \caption{
    {\bf Notation: corresponding points and outliers}. Red and blue points
    represent the discretized domains of source and target functions, respectively.
    The variable $c_n\in\{0,1\}$ indicates whether or not $x_n$ is a non-outlier.
    The variable $e_n\in\{1,\cdots,M\}$ specifies the source domain point that corresponds to $x_n$.
    Target domain points $x_1$, $x_2$, and $x_3$ represent, respectively, a point
    corresponding to $y_m$, a non-outlier not corresponding to $y_m$, and an outlier.
  }
  \label{fig:corresp}
\end{figure}

Fig. \ref{fig:corresp} explains the notation regarding corresponding points.
Fig. \ref{fig:meth:gmdl} illustrates the generative model. Hereafter, we define
the generative model on the basis of the assumptions.

\subsubsection{Domain Transformation Model}
Let us begin with Assumption 1. We define the map $\mathcal{T}$ that combines a
similarity transformation $T$ and a nonlinear displacement field
$v_Y(\cdot):\mathbb{R}^D\rightarrow\mathbb{R}^D$ as follows:
\begin{align}
   \mathcal{T}(y_m)=T(y_m+v_Y(y_m))=sR(y_m+v_m)+t, \nonumber
\end{align}
where $s>0$ is a scale factor, $R\in\mathbb{R}^{D\times D}$ is a
rotation matrix, $t\in\mathbb{R}^{D}$ is a translation vector, and
$v_m=v_Y(y_m)\in\mathbb{R}^{D}$ is the displacement vector associated with $y_m$. The
transformation model allows for rigid and nonrigid function registration in a
single algorithm. We assume $\xi=(s,R,t)$ follows a Dirac delta distribution for
simplicity. The next section defines the generative model of $v_Y(\cdot)$.

\subsubsection{Motion Coherence Prior}

\begin{figure*}
  \centering
  \includegraphics[width=0.99\textwidth]{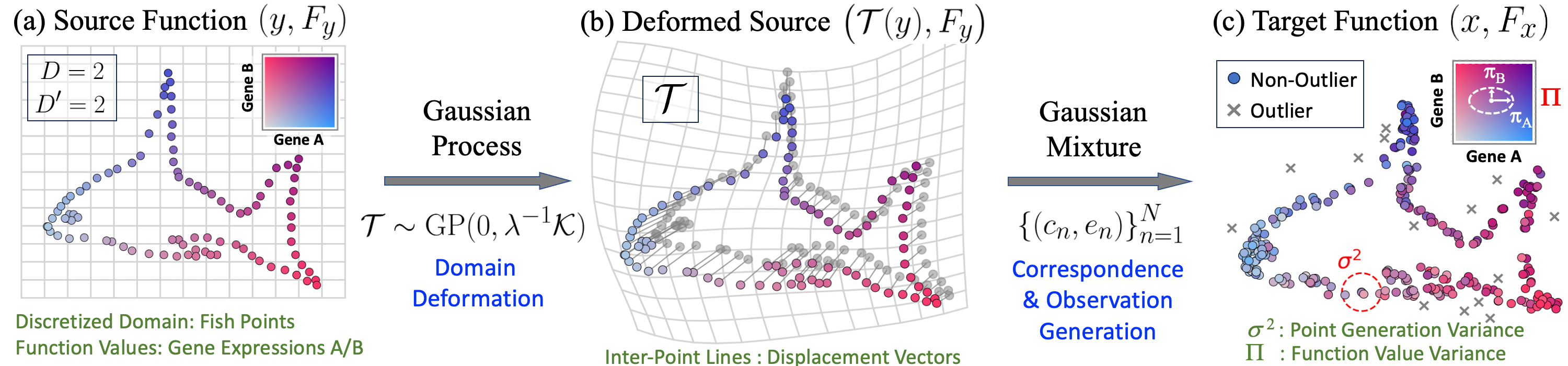} 
  \caption{
    {\bf Illustration of the generative model}. Function registration reverses this process.
    (a) Discretized source function. Point colors indicate function values.
    (b) Discretized source function deformed by a motion coherence prior, i.e., a Gaussian process.
    (c) Discretized target function generated by a mixture model. 
  }
  \label{fig:meth:gmdl}
\end{figure*}

We assume a motion coherence prior \cite{Hirose21,Hirose21a,Hirose23} generates a
displacement field $v_Y(\cdot)$, facilitating smooth domain deformation. It
independently generates a component function
$v_Y^{(d)}(z):\mathbb{R}^D\rightarrow\mathbb{R}$ for $d\in\{1,\cdots,D\}$ on the
basis of the following distribution, called a Gaussian process (GP)
\cite{Rasmussen2006}:
\begin{align}
  v_Y^{(d)}(z) \sim \text{GP}(0, \lambda^{-1}\mathcal{K}(z,z')), \nonumber
\end{align}
where $\lambda>0$ is the constant controlling the variance of $v_Y^{(d)}$, and
$\mathcal{K}:\mathbb{R}^D\times \mathbb{R}^D\rightarrow\mathbb{R}$ is a kernel
function. This implies that a discretized displacement field
$v=(v_1^T,\cdots,v_M^T)^T$ with $v_m=v_Y(y_m)$ follows the Gaussian distribution:
\begin{align}
  p(v|y) =\phi(v; 0, \lambda^{-1}G\otimes I_D), \nonumber
\end{align}
where $G=(\mathcal{K}(y_m,y_{m'}))\in \mathbb{R}^{M \times M}$ is a symmetric
matrix called the coherence matrix, and $\otimes$ is the Kronecker product.
The motion coherence prior induces smoothness in the displacement field, i.e.,
displacement vectors become increasingly parallel with decreasing distance
between points, as shown in Fig. \ref{fig:meth:gmdl}b. When registering
functions, the smoothness assumption effectively reduces the search space of $v$
because rough displacement fields have lower prior probability.
From a statistical mechanics point of view, the coherence prior can be interpreted
as an elastic force field, called the Gaussian network model \cite{Bahar1997,Haliloglu1997}.

\subsubsection{Mixture Model}
We define a core component of the generative model using a mixture model.
First, under Assumptions 5 and 6, we express the generative relation for
$(x_n,f_X(x_n))$ given $(c_n,e_n)=(1,m)$ using the following model factor:
\begin{equation}
  \phi_{mn}
  =
  \phi(x_n;\mathcal{T}(y_m),\sigma^2 I_D)
  \phi(f_X(x_n);f_Y(y_m),\Pi)^\zeta,
  \label{eq:joint_likelihood}
\end{equation}
where $\phi_{mn}$ denotes this model factor and $\zeta\geq0$ is a
parameter that balances spatial and functional information.
Strictly, when $\zeta\neq1$, the powered functional Gaussian is not
normalized as a probability density with respect to $f_X(x_n)$.
Thus, $\phi_{mn}$ is used as a tempered likelihood factor, and the
complete construction defines an unnormalized generalized probabilistic
model rather than an ordinary joint probability distribution.
For notational simplicity, we continue to use $p$ for the resulting
generalized model expressions, although they need not integrate to one
over the observations. For fixed observations, the corresponding
generalized posterior is defined, when normalizable, by normalization
with respect to $\theta$.
Throughout this section, the
term ``generative model'' refers to the directional interpretation of
the component construction.
Under this generative interpretation, $x_n$ and $f_X(x_n)$ are generated
around $\mathcal{T}(y_m)$ and $f_Y(y_m)$ under the covariance matrices
$\sigma^2 I_D$ and $\Pi$, respectively.

When registering functions, the first normal distribution evaluates the proximity between $x_n$
and $\mathcal{T}(y_m)$, and the second evaluates the similarity between
$f_X(x_n)$ and $f_Y(y_m)$, weighted by $\zeta$. This parameter acts as a
semantic coupling coefficient, determining how much the functional signal, e.g.,
gene expression, should drive the deformation of the physical domain;
Eq. (\ref{eq:joint_likelihood}) can be interpreted as a tempered, dimension-balanced likelihood.
We set $\zeta$ so that the functional term scales appropriately with feature dimension 
and remains comparable to the spatial likelihood.

Next, under Assumptions 2-6, we define the generative model of
$(x_n, f_X(x_n), e_n, c_n)$ as a nested mixture model:
\begin{multline}
p(x_n, f_X(x_n), c_n, e_n | y, F_y, v, \alpha, \xi, \sigma^2, \Pi) \\
= \{\omega p_\text{out}^{(n)}\}^{1-c_n} \left\{ (1-\omega) \prod_{m=1}^{M} (\alpha_m \phi_{mn})^{\delta_m(e_n)} \right\}^{c_n},
  \label{eq:generative_model} 
\end{multline}
where $p_\text{out}^{(n)}$ is an abbreviation of $p_\text{out}(x_n, f_X(x_n))$,
and $\delta_m(e_n)$ is the indicator function, with a value of 1 if $e_n = m$ and
0 otherwise. When registering functions, 
this model evaluates the probability that $x_n$ is a non-outlier and, for each $m$, the probability that $x_n$ corresponds to $y_m$.

\subsubsection {Mixing Probabilities}
We define a generative model for the mixing probabilities $\alpha=(\alpha_m)_{m=1}^M$,
subject to $\sum_{m=1}^M \alpha_m=1$. We assume that $\alpha$ follows the Dirichlet distribution: 
\begin{align}
  p(\alpha) =\text{Dir}(\alpha|\kappa 1_M), \nonumber
\end{align}
where $\kappa >0$ is the parameter controlling the variance of $\alpha_m$, and
$1_M$ is the vector of all 1s of size $M$. When performing function registration,
this distribution helps prevent $\alpha_m$ from becoming degenerate and stabilizes the
computation. We note that $\alpha_m$ becomes $1/M$ for all $m$ 
as $\kappa\rightarrow\infty$.

\subsubsection{Generalized Joint Model}
Combining the preceding generative components, we define the
following unnormalized generalized joint model:
\begin{align}
  &p(x,y,F_x,F_y,\theta) \nonumber \\
  &\propto  p(v|y)p(\alpha)\prod_{n=1}^{N} p(x_n,f_X(x_n),c_n,e_n|y,F_y,v,\alpha,\xi,\sigma^2,\Pi),
  \nonumber
\end{align}
where $\theta=(v,\alpha,c,e,\xi,\sigma^2,\Pi)$ is the set of unobserved variables
mediating the generation of $(x,F_x)$. This model defines how the source function
$(y,F_y)$ generates the target function $(x,F_x)$ with the latent variables
$\theta$. In practice, we use it to solve the inverse problem, i.e., function
registration; we estimate $\theta$ given $(x,y,F_x,F_y)$.

\subsubsection{Inference Issue}
To perform function registration, we need a reasonable
estimate of $\theta$, e.g., the mode of the corresponding
generalized posterior $p(\theta \mid x,y,F_x,F_y)$.
However, its exact computation is
intractable for large $M$ and $N$. A naive method requires computation over
all combinations of $c$ and $e$, amounting to $(M+1)^N$ combinations. We
avoid this issue by using variational inference, reviewed in the next section.

\subsection{Variational Inference} \label{sec:meth:vbay}
Variational inference reduces computational difficulties in Bayesian inference. This
section reviews the variational inference framework.

\subsubsection{Motivation}
In Bayesian inference, a set of unobserved variables $\theta$ is estimated from a set
of observations $z$. The estimate of $\theta$ is typically defined as the mode of a
posterior distribution $p(\theta|z)$ or the expectation of $\theta$ under
$p(\theta|z)$. Computing the estimate is, however, often intractable. For example,
the analytic form of the mode might be unavailable due to multimodality, or the
computational cost of the expectation might be prohibitively large due to the
discrete variables involved.

\subsubsection{Outline}
Variational inference approximates $p(\theta|z)$ with an alternative distribution
$q(\theta)$ whose mode or expectation is easily computable. Because $q(\theta)$
itself is generally unknown, the problem of estimating $\theta$ is replaced by
finding the approximate distribution $q(\theta)$. Typically, variational inference is
defined as the minimization of the Kullback--Leibler (KL) divergence between
$q(\theta)$ and $p(\theta|z)$.

\subsubsection{Constraints}
If no constraint is imposed on $q(\theta)$, computing the expectation and the mode
remains intractable. This is because the KL divergence is minimized when $q(\theta) =
p(\theta|z)$, returning us to the original intractable problem.
Variational inference reduces the computational difficulty by constraining
$q(\theta)$ to be the product of its marginal distributions, i.e.,
\begin{align}
 q(\theta)=\prod_{j=1}^{J} q_j(\theta_j), \nonumber
\end{align}
where $\theta_j$ is the $j$th group of $\theta=(\theta_1,\cdots,\theta_J)$ and
$q_j(\theta_j)$ is the marginal distribution of $q(\theta)$ with respect to $\theta_j$.
This factorization splits the original problem into subproblems and reduces the
computational difficulty.

\subsubsection{Procedure}
If we assume that only $q_i$ is unknown among the factorized distributions
$\{q_j\}_{j=1}^J$, the following substitution is known not to increase the KL divergence
between $q(\theta)$ and $p(\theta|z)$:
\begin{align}
  \hat{q}_i(\theta_i) \propto \exp (E_{i}[\ln p(z,\theta)]),
\label{eq:vb_solution}
\end{align}
where $E_{i}[\ln p(z,\theta)] = \int \ln p(z,\theta)\prod_{j (\neq i)}^{J}
q_j(\theta_j) d\theta_{j}$. Therefore, an approximate $\hat{q}(\theta)$ can be
obtained as follows:
\begin{enumerate}
\item Initialize $q_i$ for all $i\in\{1,\cdots,J\}$.
\item For each $i$, update $q_i$ while holding all other $q_{j}$ fixed using Eq.  (\ref{eq:vb_solution}). \label{item:cycle}
\item Repeat step \ref{item:cycle} until convergence.
\end{enumerate}
Each exact coordinate update does not increase the KL divergence \cite{Bishop06}.

\subsection{Algorithm} \label{sec:meth:algo}

We derive DET using variational inference.
We suppose $q(\theta)$ approximates $p(\theta|x,y,F_x,F_y)$
and $q(\theta)$ is the product of its marginal distributions as follows:
\begin{align}
  q(\theta) = q_1(v,\alpha)q_2(c,e)q_3(\xi,\sigma^2,\Pi).  \nonumber
\end{align}
Furthermore, we assume $q_3(\xi,\sigma^2,\Pi)$ is a Dirac delta distribution,
having a point mass at $(\xi_*,\sigma^2_*,\Pi_*)$.
For brevity, we do not distinguish random variables $(\xi,\sigma^2,\Pi)$ 
and their point estimates $(\xi_*,\sigma^2_*,\Pi_*)$ throughout this manuscript.
Accordingly, DET can be viewed as a variational-EM procedure:
$q_1$ and $q_2$ approximate the latent-variable posterior, whereas
$\xi$, $\sigma^2$, and $\Pi$ are point-estimated by optimizing the
same variational objective.

\subsubsection{Notation} \label{sec:meth:algo:notation}
Here, we list the useful symbols for describing the closed-form expressions for $\hat{q}_1$,
$\hat{q}_2$, and $\hat{q}_3$ as follows:
\begin{itemize}
\item  $G=(\mathcal{K}(y_m,y_{m'}))\in\mathbb{R}^{M\times M}$ -- the matrix
       defining the motion coherence prior, where $\mathcal{K}(\cdot,\cdot)$ is a kernel function.
\item  $P=(p_{mn}) \in [0,1]^{M\times N}$ -- the matching probability matrix, where $p_{mn}=E[c_n \delta_m(e_n)]$
       is the posterior probability of $x_n$ corresponding to $y_m$.
\item  $\nu=(\nu_1,\cdots,\nu_M)^T \in \mathbb{R}^M$ with $\nu_m = \sum_{n=1}^N p_{mn}$ -- 
       the vector of non-negative values, where $\nu_m$ is
       the estimated number of target points matching $y_m$.
\item  $\nu'=(\nu_1',\cdots,\nu_N')^T \in [0,1]^N$ with $\nu_n' = \sum_{m=1}^M p_{mn}$ -- the vector of
       probabilities, where $\nu_n'$ is the posterior probability of $x_n$ being a non-outlier.
\item  $\hat{N}=\sum_{n=1}^N \sum_{m=1}^M p_{mn} \leq N$ --  the estimated number of matching points 
       across $\{x_n\}_{n=1}^N$ and $\{y_m\}_{m=1}^M$.
\item  $\text{Tr}(\cdot),|\cdot|,\text{d}(\cdot)$ -- the trace of a matrix, the determinant of a matrix,
       and the operation converting a vector into the corresponding diagonal matrix, respectively.
\end{itemize}

We simplify the notation regarding the Kronecker product by attaching
the tilde to a matrix or a vector as follows: 
\begin{align}
  \tilde{P}=P\otimes I_D, ~\tilde{\nu}=\nu\otimes 1_D, ~\tilde{\nu}'=\nu'\otimes 1_D. \nonumber
\end{align}
As with the above notation, we denote the augmented form of the similarity transformation 
by $\tilde{T}$, i.e.,
\begin{align}
\tilde{T}(y) = s(I_M\otimes R) y +(1_M\otimes t). \nonumber
\end{align}
We summarize the closed-form expressions for $\hat{q}_1$, $\hat{q}_2$
and $\hat{q}_3$ and show how DET updates $q_1$, $q_2$, and $q_3$.
Detailed derivations are given in Appendices A–C.

\subsubsection{Local Alignment Update: $q_1(v,\alpha)$ }
The update of $q_1(v,\alpha)$ improves local registration by changing the domain deformation.
Given $q_2(c,e)$ and $q_3(\xi,\sigma^2,\Pi)$, 
we obtain the following closed-form expression for $\hat{q}_1(v,\alpha)$:

\hangindent=1.5em
\hangafter=1
  \vspace{1mm}\noindent{\bf Proposition 1.}
  {\em 
    The approximate posterior distribution $\hat{q}_1(v,\alpha)$ is factorized into its marginals, 
    i.e., $\hat{q}_1(v,\alpha)=\hat{q}_\alpha(\alpha)\hat{q}_v(v)$. 
    Furthermore, $\hat{q}_\alpha(\alpha)$ and $\hat{q}_v(v)$ are Dirichlet and Gaussian distributions,
    respectively, which are derived as follows:
  }
  \begin{align}
     \hat{q}_\alpha(\alpha)& = \text{Dir}(\alpha|\kappa 1_M +\nu), \nonumber   \\
     \hat{q}_v (v)         & = \phi\Big(v;\frac{s^2}{\sigma^2}\tilde{\Sigma} \text{d}(\tilde{\nu})
                                 (\tilde{T}^{-1}(\hat{x})-y), \tilde{\Sigma}\Big), \nonumber
  \end{align}
\hangindent=1.5em
\hangafter=1
  {\em
    where $\tilde{T}^{-1}(\hat{x})$ with $\hat{x}=\textnormal{d}(\tilde{\nu})^{-1} \tilde{P} x\in\mathbb{R}^{DM}$
    is the domain of $f_X$ that is inversely aligned from $x$ to $y$,
    and $\tilde{\Sigma}=\Sigma \otimes I_D\in\mathbb{R}^{DM\times DM}$
    with $\Sigma=(\lambda G^{-1} +\frac{s^2}{\sigma^2}\text{d}(\nu))^{-1}\in\mathbb{R}^{M\times M}$
    is the posterior covariance matrix of a displacement field $v$.
  } 

\vspace{2mm} 
This proposition suggests the posterior mean of $v$ is a kernel smoothing 
of residual vectors $\{T^{-1}(\hat{x}_m)-y_m\}_{m=1}^M$.
For convenience, we define the vectors of size $DM$ that can be decomposed into $M$ vectors of size $D$ as follows:
\begin{center}
\begin{tabular}{l}
  $\displaystyle \hat{x}=(\hat{x}_1^T,\cdots,\hat{x}_M^T)^T=\text{d}(\tilde{\nu})^{-1}\tilde{P}x$, \\
  $\displaystyle \hat{v}=(\hat{v}_1^T,\cdots,\hat{v}_M^T)^T=\frac{s^2}{\sigma^2}\tilde{\Sigma} \text{d}(\tilde{\nu})(\tilde{T}^{-1}(\hat{x})-y)$, \\
  $\displaystyle \hat{u}=(\hat{u}_1^T,\cdots,\hat{u}_M^T)^T=y+\hat{v}$, \vspace{1mm}\\
  $\displaystyle \hat{y}=(\hat{y}_1^T,\cdots,\hat{y}_M^T)^T=\tilde{T}(y+\hat{v})$. \\
\end{tabular}
\end{center}
Furthermore, the proposition suggests how DET updates
$\braket{\alpha_m}=\exp(E[\ln\alpha_m])$ and $\braket{\phi_{mn}}=\exp(E[\ln\phi_{mn}])$:
\begin{align}
    \textstyle \braket{\alpha_m}  &\textstyle = \exp\big\{\psi(\kappa+\nu_m)-\psi(\kappa M +\hat{N}) \big\}, \nonumber \\
    \braket{\phi_{mn}}            &\textstyle = b_m \phi(x_n;\hat{y}_m,\sigma^2 I_D) \phi(f_X(x_n);f_Y(y_m),\Pi)^{\zeta}, \nonumber
\end{align}
where $\psi(\cdot)$ is the digamma function,
$b_m$ is defined as $\exp\{-\frac{s^2}{2\sigma^2}\text{Tr}(\sigma_m^2 I_D)\}$, and
$\sigma^2_m$ is the $m$th diagonal element of $\Sigma$.
These terms are required for updating $q_2(c,e)$.
See Appendices D and E for their derivations.

\subsubsection{Correspondence Update: $q_2(c,e)$ }
The update of $q_2(c,e)$ improves the matching probability matrix $P=(p_{mn})$.
Given $q_1(v,\alpha)$ and $q_3(\xi,\sigma^2,\Pi)$,
we obtain the following closed-form expression for $\hat{q}_2(c,e)$:

\hangindent=1.5em
\hangafter=1
\vspace{1mm}\noindent{\bf Proposition 2.}
{\em 
  The approximate posterior distribution $\hat{q}_2(c,e)$ is a combination of a Bernoulli distribution
  and a categorical distribution, derived as follows:
}
\begin{align}
  \hat{q}_2 (c,e) &= \prod_{n=1}^{N}
                         (1-{\nu_{n}'} )^{1-c_n}
                       \bigg\{
                            {\nu_{n}'}
                            \prod_{m=1}^M
                              \Big(\frac{p_{mn}}{\nu_n'}\Big)^{\delta_m(e_n)}
                       \bigg\}^{c_n}, \nonumber
\end{align}
{\em
  where $p_{mn}$ and $\nu_n'$ are defined as
}
\begin{align}
  p_{mn}=\frac{(1-\omega)  \braket{\alpha_m}\braket{\phi_{mn}}  }  
              {   \omega   p_\text{out}^{(n)}+(1-\omega) \sum _{m'=1}^M  \braket{\alpha_{m'}}\braket{\phi_{m'n}}},
  \nonumber 
\end{align}
{\em and $\nu_n'=\sum_{m=1}^M p_{mn}$.}

\vspace{2mm}
This proposition suggests how DET updates $P$ and other related terms
such as $\nu$, $\nu'$, and $\hat{N}$.
It also suggests that $\hat{q}_2(c,e)$ is factorized into 
$\prod_{n=1}^N \hat{q}_2^{(n)}(c_n,e_n)$. Furthermore, the proposition provides
the following observations, which are consistent with the description in Section \ref{sec:meth:algo:notation}:
\begin{itemize}
  \item The definition of $p_{mn}$ is consistent with the proposition
        because $E[c_n \delta_m(e_n)]=\hat{q}_2^{(n)}(c_n=1,e_n=m)=p_{mn}$.
  \item The posterior marginal distribution of $c_n$ is the Bernoulli distribution with
        the probability $\nu_n'$, and thus, the posterior probability of $x_n$ being a non-outlier is $\nu_n'$.
  \item The number of target points matching $y_m$ can be estimated as $\nu_m$
        because $E[\sum_{n=1}^N c_n\delta_m(e_n)]=\nu_m$.
  \item The number of matching points between the target and source point sets, $x$ and $y$,
        can be estimated as $\hat{N}$ because $E[\sum_{n=1}^N \sum_{m=1}^M c_n\delta_m(e_n)]=\hat{N}$.
\end{itemize}

\subsubsection{Global Alignment Update: $q_3(\xi,\sigma^2,\Pi)$}
The update of $q_3$ improves the global registration defined by $\xi=(s,R,t)$.
Because $q_3$ is used only as a compact representation of point
estimates, we directly optimize the variational objective with
respect to $(\xi,\sigma^2,\Pi)$ given $q_1(v,\alpha)$ and
$q_2(c,e)$, rather than applying Eq.~(\ref{eq:vb_solution}).
Let us define the following notation:
\begin{center}
\begin{tabular}{l}
  $\displaystyle (\bar{x}^T,\bar{u}^T,\bar{\sigma}^2) =\frac{1}{\hat{N}}\sum_{m=1}^M \nu_m (\hat{x}_m^T,\hat{u}_m^T,\sigma_m^2)$,~ \vspace{.5mm} \\
  $S_{xu}= \displaystyle \frac{1}{\hat{N}}\sum_{m=1}^M \nu_m (\hat{x}_m-\bar{x})(\hat{u}_m-\bar{u})^T$,                      \vspace{1.5mm} \\ 
  $S_{uu}= \displaystyle \frac{1}{\hat{N}}\sum_{m=1}^M \nu_m (\hat{u}_m-\bar{u})(\hat{u}_m-\bar{u})^T  +\bar{\sigma}^2 I_D$, \vspace{0mm} \\
\end{tabular}
\end{center}
where $\hat{x}_m\in\mathbb{R}^D$ and $\hat{u}_m\in\mathbb{R}^D$ are the $m$th subvectors of $\hat{x}$ and $\hat{u}$,
and $\sigma_m^2$ is the $m$th diagonal element of $\Sigma$.
With this notation, we obtain the following proposition:

\hangindent=1.5em
\hangafter=1
\vspace{1mm}\noindent{\bf Proposition 3.}
{\em
 Assume $\widehat{N}>0$ and $\zeta>0$. Given $q_1$ and $q_2$,
 the point-parameter block $(\xi,\sigma^2,\Pi)$ is updated by
 the following formulae:
}

\vspace{1mm}
{\setlength{\tabcolsep}{3pt}
\begin{tabular}{cl}
  $\hat{R}$        &\hspace{-4mm} $\displaystyle = \Phi \text{d}(1,\cdots,1,|\Phi\Psi^T|)\Psi^T$,                        \vspace{0.5mm}\\
  $\hat{s}$        &\hspace{-4mm} $\displaystyle = \text{Tr}\big( \hat{R}^T S_{xu}\big)\big/ \text{Tr}\big(S_{uu}\big)$, \vspace{0.5mm}\\
  $\hat{t}$        &\hspace{-4mm} $\displaystyle = \bar{x}- \hat{s}\hat{R}\bar{u}$,                                      \vspace{0.5mm}\\
  $\hat{\sigma}^2$ &\hspace{-4mm} $\displaystyle =
    \frac{1}{\hat{N}D} \big\{x^T \text{d}({\tilde{\nu}'}) x-2\hat{y}^T\tilde{P}x + \hat{y}^T\text{d}(\tilde{\nu})\hat{y}\big\}
                            +\hat{s}^2 \bar{\sigma}^2$,\nonumber \vspace{0.8mm}\\
  $\hat{\Pi}$      &\hspace{-4mm} $\displaystyle =
    \frac{1}{\hat{N}}  \big\{F_x \text{d}({{\nu}'}) F_x^T -2\text{sym}(F_y{P}{F}_x^T) + {F}_y\text{d}({\nu}){F}_y^T\big\}$, \nonumber \vspace{1mm} \\
\end{tabular}}
\hangindent=1.5em
\hangafter=1
{\em
  where $\hat{y}=\tilde{T}(y+\hat{v})$ indicates the points constituting the transformed domain,
  and $\Phi$ and $\Psi$ are the orthogonal matrices of size $D\times D$
  obtained by the singular value decomposition of $S_{xu}$, i.e., $S_{xu}=\Phi \Lambda \Psi^T$
  with a diagonal matrix $\Lambda$. The function $\text{sym}(\cdot)$ is defined as $\text{sym}(A)=(A+A^T)/2$.
} 

\vspace{2mm}
We note that $\sigma^2$ and $\Pi$ indicate how loose the matching criterion is,
and their updates gradually shrink a candidate set of target points that could match
each source point \cite{Hirose21}.

\subsubsection{Initialization}
Variational inference requires initializing $q(\theta)$,
which corresponds to initializing the expected random variables. 
We initialize them in a non-informative manner;
we set $\hat{v}=0$, $\braket{\alpha_m}=1/M$, $s=1$, $R=I_D$, and $t=0$.
In addition, we initialize $\sigma^2$ and $\Pi$ using a constant $\gamma>0$ as follows:
\begin{align}
 &\sigma^2   = \frac{\gamma}{NMD}{ \sum_{n=1}^{N}\sum_{m=1}^{M} ||x_n-y_m||^2},        \nonumber \\ 
 &\Pi        = \frac{\gamma}{NM} {\sum_{n=1}^{N}\sum_{m=1}^{M} Q(f_X(x_n)-f_Y(y_m))}, \nonumber 
\end{align}
where $Q(\cdot)$ is the function defined as $Q(a)=aa^T$.
The parameter $\gamma$ controls the granularity of registration \cite{Hirose23}.
A small $\gamma$ is suitable for fine registration and is typically 
used for hierarchical registration described in Section \ref{sec:impl:hopt}.

\begin{figure*}
  \includegraphics[width=\textwidth]{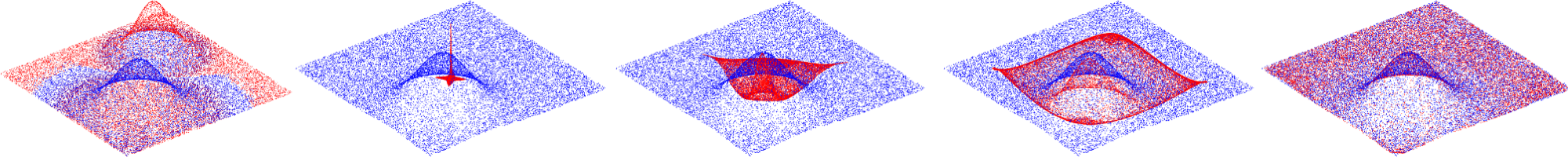}
  \caption{  
     {\bf Function registration using the DET algorithm}.
     The target and source functions are shown in the leftmost figure
     and colored blue and red, respectively. Registration proceeds from left to right.
  } \label{fig:demo:2d}
\end{figure*}

\begin{figure}
  \vspace{-5mm}
  \begin{tabular}{l} \\ \hline
    {\bf Algorithm 1}: {\bf Domain Elastic Transform}     \\ \hline \vspace{-3mm}\\
    \vspace{1mm} {\bf Input}:  ~$\omega\in[0,1),~\lambda>0,~\kappa>0,~\gamma>0,~\eta>0,~\mathcal{K}(\cdot,\cdot)$\\
    \vspace{1mm}\hspace{.3cm}    $x=(x_1^T,\cdots, x_N^T)^T\in\mathbb{R}^{DN}$ \\
    \vspace{1mm}\hspace{.3cm}    $y=(y_1^T,\cdots, y_M^T)^T\in\mathbb{R}^{DM}$ \\
    \vspace{1mm}\hspace{.3cm}    $F_x=(f_X(x_1),\cdots,f_X(x_N))\in\mathbb{R}^{D'\times N}$ \\
    \vspace{1mm}\hspace{.3cm}    $F_y=(f_Y(y_1),\cdots,f_Y(y_M))\in\mathbb{R}^{D'\times M}$ \\
    \vspace{1.4mm} {\bf Output}: $ \hat{y}=\tilde{T}(y+\hat{v})=s(I_M\otimes R)(y+\hat{v})+(1_M\otimes t)$ \\ \hline \vspace{-3mm}\\
    \vspace{1.4mm} {\bf Initialization}: \\
    \hspace{0.3cm}\vspace{1.4mm}   $\hat{y}=y$, ~$\hat{v}=0$, ~$\Sigma=I_{M}$, ~$s=1$, ~$R=I_D$, ~$t=0$ \\
    \hspace{0.3cm}\vspace{1.4mm}   $G=(\mathcal{K}(y_m,y_{m'}))$, ~$\braket{\alpha_m}=\frac{1}{M}$, ~$\zeta=\eta \frac{D}{D'}$\\
    \hspace{0.3cm}\vspace{1.4mm}   $\sigma^2  =\frac{\gamma}{N M D} \sum_{n=1}^N \sum_{m=1}^M ||x_n-y_m||^2$ \\
    \hspace{0.3cm}\vspace{2.5mm}   $\Pi       =\frac{\gamma}{N M}   \sum_{n=1}^N \sum_{m=1}^M Q(f_X(x_n)-f_Y(y_m))$ \\
    \vspace{2.0mm}  {\bf Optimization}: Repeat a), b), and c) until convergence. \\
    \hspace{0.0cm}\vspace{1.5mm} a) Update matching $P=(p_{mn})$ and related terms.\\
    \hspace{0.3cm}\vspace{1.4mm}   $b_m=\exp\{-\frac{s^2}{2\sigma^2}\text{Tr}(\sigma_m^2 I_D)\}$ \\
    \hspace{0.3cm}\vspace{1.4mm}   $\braket{\phi_{mn}}= b_m \phi(x_n;\hat{y}_m,\sigma^2 I_D) \phi(f_X(x_n);f_Y(y_m),\Pi)^{\zeta}$ \\
    \hspace{0.3cm}\vspace{1.4mm}   $p_{mn}=
                                    \frac{(1-\omega)\braket{\alpha_m}\braket{\phi_{mn}}}
                                         {\omega p_\text{out}(x_n,f_X(x_n)) +(1-\omega)\sum_{m'=1} \braket{\alpha_{m'}}\braket{\phi_{m'n}}}$ \\
    \hspace{0.3cm}\vspace{2.5mm}   $\nu=P1_N$, ~$\nu'=P^T1_M$, ~$\hat{N}=\nu^T 1_M$, ~
                                   $\hat{x}=\text{d}(\tilde{\nu})^{-1}\tilde{P}x$         \\
    \hspace{0.0cm}\vspace{1.5mm} b) Update local alignment $\hat{v}$ and related terms.\\
    \hspace{0.3cm}\vspace{1.4mm}   $\Sigma^{-1}=\lambda G^{-1} +\frac{s^2}{\sigma^2} \text{d}(\nu) $, ~
                                   $\hat{v}=\frac{s^2}{\sigma^2}\tilde{\Sigma}\text{d}(\tilde{\nu})\big(\tilde{T}^{-1}(\hat{x})-y \big)$ \\
    \hspace{0.3cm}\vspace{2.5mm}   $\hat{u}=y+\hat{v}$, ~$\braket{\alpha_m} = \exp\big\{\psi(\kappa+\nu_m) - \psi\big(\kappa M+\hat{N} \big) \big\}$ \\
    \hspace{0.0cm}\vspace{1.5mm} c) Update global alignment $(s,R,t)$ and related terms. \\
    \hspace{0.3cm}\vspace{1.4mm}   $(\bar{x}^T,\bar{u}^T,\bar{\sigma}^2)^T =\frac{1}{\hat{N}}\sum_{m=1}^M \nu_m (\hat{x}_m^T,\hat{u}_m^T,\sigma_m^2)^T$ \\
    \hspace{0.3cm}\vspace{1.4mm}   $S_{xu} =\frac{1}{\hat{N}} \sum_{m=1}^M \nu_m (\hat{x}_m-\bar{x})(\hat{u}_m-\bar{u})^T$ \\
    \hspace{0.3cm}\vspace{2.0mm}   $S_{uu} =\frac{1}{\hat{N}} \sum_{m=1}^M \nu_m (\hat{u}_m-\bar{u})(\hat{u}_m-\bar{u})^T+\bar{\sigma}^2 I_D$ \\
    \hspace{0.3cm}\vspace{1.4mm}   $\Phi \Lambda\Psi^T=\text{svd}(S_{xu})$, ~$R=\Phi\text{d}(1,\cdots,1,|\Phi\Psi^T|)\Psi^T$   \\
    \hspace{0.3cm}\vspace{1.4mm}   $s=\text{Tr}(R^TS_{xu})/\text{Tr}(S_{uu})$, ~$t=\bar{x}-sR\bar{u}$, ~$\hat{y}=\tilde{T}(y+\hat{v})$ \\
    \hspace{0.3cm}\vspace{1.4mm}   $\sigma^2=\frac{1}{\hat{N}D}
                                     \big\{x^T\text{d}({\tilde{\nu}}')x -2\hat{y}^T\tilde{P}x + \hat{y}^T\text{d}(\tilde{\nu})\hat{y}\big\}
                                     +s^2\bar{\sigma}^2$  \\
    \hspace{0.3cm}\vspace{1.4mm}   $\Pi=\frac{1}{\hat{N}}
                                      \big\{{F}_x\text{d}({{\nu}}'){F}_x^T -2\text{sym}({F}_y{P}{F}_x^T) + {F}_y\text{d}({\nu}){F}_y^T\big\}$  \\ 
                                   \hline\\
  \end{tabular}
  {\footnotesize
    A tilde symbol represents the Kronecker product following the rules:
    $\tilde{A}=A\otimes I_D$ and $\tilde{a}=a\otimes 1_D$, where $A$ is a matrix and $a$ is a vector.
    The symbols $\phi$ and $\psi$ represent Gaussian and digamma functions.
    The $m$th subvectors of $\hat{x}$ and $\hat{u}$ are represented as $\hat{x}_m$ and $\hat{u}_m$, respectively.
    The $m$th diagonal element of $\Sigma$ is denoted by $\sigma_m^2$.
    $\text{svd}(A)$ represents singular value decomposition; $Q(a)=aa^T$; $\text{sym}(A)=(A+A^T)/2$. 
    $|\cdot|$ and $\text{Tr}(\cdot)$ are the determinant and trace of a matrix,
    and $\text{d}(\cdot)$ is the operation converting a vector into its diagonal matrix.
  }
\end{figure}

\subsubsection{Summary and Comments}
We summarize DET in Algorithm 1. 
Fig.~\ref{fig:demo:2d} shows how the algorithm proceeds when applied to Gaussian functions.
Supplementary Video 1 shows the corresponding complete registration trajectory.
We note that DET performs rigid function registration under $(\lambda,s)=(\infty,1)$ or $(v,s)=(0,1)$.
We also note that DET extends BCPD from geometric point-set registration to high-dimensional function
registration by introducing a spatial-functional likelihood and the enhancement components described
in Section \ref{sec:impl}. Therefore, BCPD is recovered as a degenerate special case, i.e., $\zeta=0$,
where the functional likelihood is removed and only the spatial likelihood remains.
The variational-inference machinery, motion-coherence prior, and acceleration strategy are
inherited from BCPD~\cite{Hirose21,Hirose21a,Hirose23}.

\section{Additional Techniques} \label{sec:impl}
Section~\ref{sc:bcpd} presented the probabilistic formulation and inference
algorithm of DET. This section describes additional techniques for practical
high-dimensional function registration. We distinguish core components from
optional components. 

\subsection{Core Implementation}
\subsubsection{Likelihood Balancing} \label{sec:impl:like}

To ensure robustness across varying feature dimensions, we decompose the weighting
parameter $\zeta$ into a confidence coefficient and a dimensionality-balancing factor:
\begin{equation}
\zeta = \eta (D/D').
\label{eq:zeta_balancing}
\end{equation}
The ratio $D/D'$ compensates for the difference between the accumulated spatial
and functional energy scales, preventing high-dimensional functional signals
from overwhelming geometric information.
The parameter $\eta> 0$ represents the \textit{relative confidence}
in the functional signal. We set $\eta = 1.0$ as the default, assuming equal
reliability between geometry and function.

\subsubsection{Functional Annealing}
\label{sec:impl:hopt}

We extend DET to support hierarchical registration \cite{Hirose23}. DET requires predefined
parameters $\Theta$, listed in Table \ref{tab:param}. Because these parameters remain fixed
until convergence, DET may fail to register functions with large differences under weak coherence
or those with subtle differences under strong coherence. To mitigate such issues, we register
functions in a coarse-to-fine manner, i.e., by repeatedly applying DET while gradually reducing motion coherence
and adjusting functional confidence. We define hierarchical registration as follows:
\begin{equation}
\hat{y}^{(0)} = y, \quad \hat{y}^{(l)} = \mathcal{S}(x, \hat{y}^{(l-1)}, F_x, F_y, \Theta^{(l)}),
\label{eq:functional_annealing}
\end{equation}
where $l>0$ is the level of the hierarchy, and $\mathcal{S}$ represents the DET algorithm
that returns a deformed domain $\hat{y}^{(l)}$. 

Strategically, we design the sequence $\{\Theta^{(l)}\}_{l=1}^{L}$ to anneal both the spatial rigidity
and the functional weight. In the early stages, we set a large kernel width $\beta$
and a higher functional confidence $\eta$. This allows the algorithm 
to utilize global functional signals to resolve large-scale ambiguities,
such as incorrect rotation or initial misalignment.
In later stages, we reduce $\beta$ to capture fine details and decrease $\eta$. 
This \textit{functional annealing} strategy shifts the optimization priority toward spatial fidelity in the final steps,
ensuring that the registered points adhere precisely to the geometric surface of the target manifold,
rather than drifting to match noisy functional signals.

\subsubsection{Tractable Functional Covariance}

Although Proposition 3 and Algorithm 1 give an unconstrained covariance formulation for $\Pi$,
estimating a full $D' \times D'$ covariance matrix from
uncertain soft correspondences can be computationally expensive and
ill-conditioned in high-dimensional settings. In the implementation used
in our experiments, we therefore restrict $\Pi$ to a diagonal form:
\[
\pi_d^2
=
\frac{1}{\hat{N}}
\sum_{n=1}^{N}
\sum_{m=1}^{M}
p_{mn}
\left(
f_X^{(d)}(x_n) - f_Y^{(d)}(y_m)
\right)^2 .
\]

Accordingly, in our experiments, both the initialization and subsequent
updates of $\Pi$ are restricted to diagonal form; the updates use the formula above.
This parameterization reduces the number of covariance parameters from
$\mathcal{O}(D'^2)$ to $\mathcal{O}(D')$, avoids unstable cross-feature
covariance estimates during early iterations, and retains feature-wise
adaptation to unequal residual variation. The isotropic model
$\Pi = \pi^2 I_{D'}$ is a more restrictive special case that assumes equal
residual variance in every functional dimension.

\subsubsection{Acceleration}  \label{sec:impl:accl}
Acceleration is inherited from the BCPD framework rather than introduced as a
new numerical contribution here, but it is a core practical component of DET's
atlas-scale implementation. We accelerate DET using the same techniques as those applied to BCPD
\cite{Hirose21,Hirose21a,Hirose23}, as DET and BCPD share identical bottleneck
computations. Here, we outline the acceleration strategy, referring readers to the
aforementioned articles for full details.

There are two primary acceleration methods. The first method accelerates the
algorithm's internal computations, specifically the updates involving 
the matching probability matrix $P\in\mathbb{R}^{M\times N}$ and the coherence matrix $G\in\mathbb{R}^{M\times M}$.
We employ the Nystr\"om method \cite{Williams01} and k-D tree search
\cite{Bentley75} to approximate these matrices \cite{Hirose21}. The Nystr\"om
method provides a low-rank approximation, governed by the rank constraint
parameters $J$ and $K$ for $P$ and $G$, respectively. \textcolor{black}{After the early optimization stage}, we switch
from the Nystr\"om method to k-D tree search for updating $P$. This accelerates
computation by pruning low-probability correspondences, providing a significantly
more accurate approximation than the Nystr\"om method when $P$
becomes sparse. The computational and storage costs are reduced to $O((M+N)\log (M+N))$
and $O(M+N)$, respectively.

The second method accelerates the registration workflow via pre-processing and
post-processing \cite{Hirose21a,Hirose23}. This approach divides function
registration into three steps:
\begin{enumerate}
  \item Downsampling the discretized functions.
  \item Registering the downsampled discretized functions.
  \item Interpolating the domain displacement vectors to the original resolution.
\end{enumerate}
We note that the downsampling procedure requires the parameters $M'$ and $N'$,
corresponding to the number of source and target domain points after downsampling,
respectively. The computational and storage costs of this method are both $O(M+N)$.
These techniques allow DET to scale efficiently even when both $M$ and $N$ are on the order of millions.

\subsection{Optional Components}
\subsubsection{Feature-Sensitive Sampling}

While the acceleration methods \cite{Hirose21a,Hirose23} enable scalability to millions of points,
uniform downsampling can under-represent localized interfaces or rare high-variability
regions when they occupy only a small fraction of a large point cloud. In that
regime, much of a uniform landmark budget is spent on homogeneous interiors,
whereas functional discontinuities characterized by sharp changes in gene expression
may correspond to physical organ boundaries or tissue interfaces.
Geometry also plays a vital role; distinct
anatomical features, such as ventricular spaces or outer boundaries, must be preserved
even if the surrounding tissue is functionally homogeneous.

As an optional way to emphasize such localized regions, we introduce
Variance-Guided Importance Sampling (VGIS). We partition the spatial domain into
non-overlapping voxels and compute a sampling probability $p(z_i)$
using the maximum of the functional variability and geometric discontinuity:
\begin{equation}
    p(z_i) \propto \max \left(\sigma_f(z_i), \lambda_g \sigma_g(z_i) \right) + \epsilon,
    \label{eq:vgis_probability}
\end{equation}
where $\epsilon>0$ is a base constant ensuring non-zero sampling density in homogeneous regions,
and $\lambda_g>0$ is a parameter that balances the two signals.
 
The term $\sigma_f(z_i)$ captures the local functional variability, defined as the
maximum standard deviation of feature values within the voxel containing $z_i$:
\begin{equation}
    \sigma_{f}(z_i) = \max_{d} \left\{ \text{Var}_{j \in \text{voxel}(z_i)}^{1/2} [f^{(d)}(z_j)] \right\}.
    \nonumber
\end{equation}
By selecting the maximum standard deviation across feature dimensions d, this metric increases
the sampling density wherever any individual functional signal changes rapidly—for example,
at a sharp transition in a particular gene-expression marker—without allowing that signal
to be diluted by non-informative dimensions.

The term $\sigma_g(z_i)$ captures spatial boundaries, defined as the fraction of empty
face-adjacent neighbors of the voxel containing $z_i$. To prevent the sampling of isolated outliers,
we strictly set $\sigma_g(z_i) = 0$ if all neighbors are empty. 
This term becomes non-zero only at the physical edges of the manifold or boundaries of internal holes,
where the neighborhood occupancy is incomplete.

\subsubsection{Coherence Matrix}
We use the surface coherence matrix \cite{Hirose23} as an example of $G$. The
matrix is designed to combine the Gaussian kernel and the geodesic exponential
kernel, allowing for more flexible domain deformation than the Gaussian kernel.
The surface coherence matrix $G=(\mathcal{K}(y_m,y_{m'}))$ is defined as
follows:
\begin{align}
  \mathcal{K}(y_m,y_{m'}) =
                 \tau \phi_\beta(\mathcal{D}_{m{m'}}^\text{(geo)})
            +(1 -\tau)\phi_\beta(\mathcal{D}_{m{m'}}^\text{(euc)}), \nonumber
\end{align}
where $\tau\in[0,1]$ is the mixing rate, $\phi_\beta(c)=\exp(-c^2/{2\beta^2})$ is a
Gaussian function with the width parameter $\beta>0$, and
$\mathcal{D}_{m{m'}}^\text{(geo)}$ and $\mathcal{D}_{m{m'}}^\text{(euc)}$ are the
geodesic and Euclidean distances between $y_m$ and $y_{m'}$, respectively.
 
We compute $\mathcal{D}^\text{(geo)}$ on a $k$-nearest-neighbor graph constructed in
an augmented coordinate system that combines spatial locations and weighted
functional features. This construction helps topologically separate closely located
boundaries, e.g., organ interfaces. Since the geodesic kernel is generally
indefinite, we ensure the validity of the Gaussian process prior by applying the fast
positive-semidefinite approximation \cite{Hirose23}.
This effectively projects the boundary-aware kernel onto a valid covariance space,
ensuring stable deformation even across topological cuts.

\subsubsection{Adaptive Outlier Model}
A uniform functional outlier density becomes difficult to calibrate as feature
dimensionality increases because its value depends strongly on the selected
feature-space volume. DET therefore provides an optional \textit{adaptive outlier
model} that incorporates the functional weight $\zeta$ to balance the
energy scales:
\begin{equation}
p_\text{out}(z, z') = 
\begin{cases} 
  \frac{1}{V_{X}} \cdot \left(\frac{1}{V_{f_X}}\right)^{\zeta}   & \text{if } D'\leq D_0\text{,} \\
  \textcolor{black}{\frac{1}{V_{X}}} \cdot \left(\phi_{f_X}(z')\right)^{\zeta}      & \text{otherwise,}
\end{cases}
\label{eq:adaptive_outlier}
\end{equation}
where $V_{X}$ and $V_{f_X}$ are the volumes of the spatial and functional 
bounding boxes, and $\phi_{f_X}$ denotes the Gaussian density
parameterized by the target's marginal statistics. 
Like Eq.~(\ref{eq:joint_likelihood}), Eq.~(\ref{eq:adaptive_outlier}) is a generalized model factor
and is not, in general, normalized with respect to $z'$ when $\zeta\neq 1$.
Only its relative scale in the correspondence update is used.
Crucially, applying the same exponent $\zeta$ to the functional
components matches the functional scaling of the outlier and
non-outlier model factors, preventing the high-dimensional functional
term from dominating the outlier decision boundary.

The threshold $D_0=10$ is a conservative implementation cutoff for avoiding
a uniform functional background density in moderately high-dimensional feature spaces,
where uniform volumes rapidly become uninformative. Supplementary Sec. S4.6 shows
that varying this cutoff has only minor effects around the default setting,
supporting its use as a modeling cutoff rather than a dataset-specific tuning parameter.

\begin{table*}
  \caption{List of Tuning Parameters}
  \label{tab:param}
  \begin{center}
  \begin{tabular}{lcl} \hline
              & Recommended      & Description    \\ \hline
   $\lambda$  & $[1, 5000]$      & Stiffness of domain deformation; $\sqrt{\lambda^{-1}D}$ equals the root-mean-square deformation length when $\mathcal{K}(y_m,y_m)=1$.    \Tstruts \\   
   $\omega$   & $[0.0, 0.9] $    & Outlier probability. The larger, the more robust against outliers (although less sensitive to the target points).      \\ 
   $\gamma$   & $[0.1, 30]  $    & Weight given to the initial alignment; smaller values enforce it more strongly.                                        \\
   $\tau$     & $[0.0, 1.0] $    & Mixing rate between two types of motion coherence that originate from the Euclidean and geodesic distances.            \\
   $\beta$    & $[0.1, 2.5]$     & Smoothness parameter of the domain displacement field (the width of $\phi_\beta$). The larger, the smoother.           \\
   $\eta$     & $[0.5, 5.0]$     & Relative confidence in functional fidelity vs. spatial proximity (Default: 1.0).                              \Bstruts \\ \hline 
     $J$      & $[300, 600]$     & The number of points for approximating matching probabilities $P$. The smaller, the faster (but less accurate). \Tstruts \\
     $K$      & $[100, 300]$     & The number of points for approximating the coherence matrix $G$. The smaller, the faster (but less accurate).          \\
     $M'$     & $[2000, 50000]$  & The number of source points after downsampling. The smaller, the faster (but less accurate).                           \\
     $N'$     & $[2000, 50000]$  & The number of target points after downsampling. The smaller, the faster (but less accurate).                  \Bstruts \\ \hline
  \end{tabular} 
  \end{center}
  {\em 
    The first six parameters must be predefined.
    The last four parameters are acceleration parameters, and the DET algorithm runs without them.
    A nonzero $\tau$ requires $K$ to be predefined to avoid an indefinite $G$.
  }
\end{table*}

\subsection{Preprocessing and Parameters}
\subsubsection{Normalization}
We normalize input functions before registration to facilitate parameter setting and ensure the
validity of the fixed weighting parameter $\zeta = \eta(D/D')$. For arbitrary matrices
$A$ and $B$, we define the normalization operator $\mathcal{N}_B(A)$ as:
\begin{equation}
\mathcal{N}_B(A) := (A - \mu_B 1^{T})/\sigma_B,
\nonumber
\end{equation}
where $\mu_B$ is the vector containing the mean of each row in $B$, $1$ is the vector of all ones,
and $\sigma_B$ is the pooled standard deviation after feature-wise centering.
Using a scalar scale factor preserves the geometric aspect ratio of the domain.

For domain points, we use a context-aware scheme:
\begin{equation}
X' = \mathcal{N}_X(X), \quad Y' =
\begin{cases}
\mathcal{N}_X(Y) & \text{if pre-aligned,} \\
\mathcal{N}_Y(Y) & \text{otherwise,}
\end{cases}
\nonumber
\end{equation}
where $X=(x_1,\cdots,x_N)$ and $Y=(y_1,\cdots,y_M)$ are the matrix representations of the domain points $x$ and $y$.
In the unaligned case, this centers each point set independently; in the pre-aligned case,
the source is normalized using the target statistics to preserve the pre-aligned coordinate relationship.

For functional values, e.g., gene expression, we apply the same operator independently to the
target and source:
\begin{equation}
F_x' = \mathcal{N}_{F_x}(F_x), \qquad
F_y' = \mathcal{N}_{F_y}(F_y).
\nonumber
\end{equation}
Thus, each feature is centered by its dataset-specific mean, while all features in a dataset share
one scalar scale factor computed from the full functional matrix. This removes feature-specific
offsets and controls functional scale across datasets without forcing unit variance per feature.
Consequently, $\zeta$ balances aggregate spatial and functional energies while preserving relative 
variation across feature dimensions. The normalization therefore emphasizes relative spatial
expression patterns rather than absolute levels, which suits the slice- and atlas-alignment
experiments considered here because batch effects, sequencing depth, and global intensity scale
may differ among samples.


\subsubsection{Tuning Parameters} \label{sec:impl:para}

Table \ref{tab:param} summarizes the algorithm's parameters. We omitted $\kappa$
from the table because the best setting is typically $\kappa=\infty$, which
enforces $\alpha_m=1/M$ for all $m$. In this case, DET skips computing $\braket{\alpha_m}$.

\section{Experiments}\label{sec:exp}

\begin{table} 
  \caption{Parameter Settings}
  \label{tab:param:set}
  \begin{center}
  \footnotesize
  \setlength{\tabcolsep}{5pt}
  \begin{tabular}{lccccccccccc} \hline
   Sec.               & Stage & $\lambda$ & $\omega$ & $\gamma$  & $\beta$ & $\tau$ & $\eta$ & $J$ & $K$  & $M'$   & $N'$    \\ \hline
   \ref{sec:exp:merf} & L1    & -         &  0.01    &   30      &    -    &  -     &  5     &  -  &  -    & 2k     & 2k     \\
   \ref{sec:exp:merf} & L2    & 10        &  0.1     &   0.1     &    1    &  0     &  2     & 300 & 100   & 5k     & 5k     \\
   \ref{sec:exp:msta} & L1    & -         &  0.2     &   1       &    -    &  -     &  1     &  -  &  -    & 2k     & 2k     \\
   \ref{sec:exp:msta} & L2    & 1         &  0.1     &   0.1     &    1    &  0     &  1     & 300 & 150   & 10k    & 10k    \\ \hline
  \end{tabular}
  \end{center}
  {\em A hyphen (-) indicates that the corresponding parameter was not used;
   the similarity transformation does not require $\lambda$, $\beta$, $\tau$, and $K$.
   L1 and L2 in the second column indicate the similarity and nonrigid stages, respectively.
   In the L1 stage, $J$ was not used because DET is sufficiently fast with $M'=N'=2000$.
  }
\end{table}

This section evaluates the performance of DET.
Section \ref{sec:exp:merf} presents a comparative study using MERFISH data.
Section \ref{sec:exp:msta} demonstrates spatiotemporal registration on Stereo-seq atlases.
The parameters used for these experiments are shown in Table \ref{tab:param:set}.
The supplementary material provides formal metric definitions, reproducibility details,
ablation studies, sensitivity analyses, feature-dimensionality stress tests, and
DET peak-memory measurements. It also provides demonstrations of DET's versatility,
e.g., registering audio signals, images, and shapes.

\subsection{Slice-to-Slice Alignment (MERFISH)} \label{sec:exp:merf}

To evaluate DET on high-dimensional spatial transcriptomics data, we used the
Mouse Brain MERFISH Zhuang-ABCA-1 dataset~\cite{Zhang23}. The task is to
register adjacent coronal slices represented as irregular point sets, where each
cell has two-dimensional centroid coordinates and a high-dimensional expression profile.
This setting is challenging because the tissue geometry is sparse and nonrigid,
while the functional signal is noisy but essential for resolving ambiguous
correspondences.

\begin{figure*}[t]
  \centering
  \includegraphics[width=0.98\textwidth]{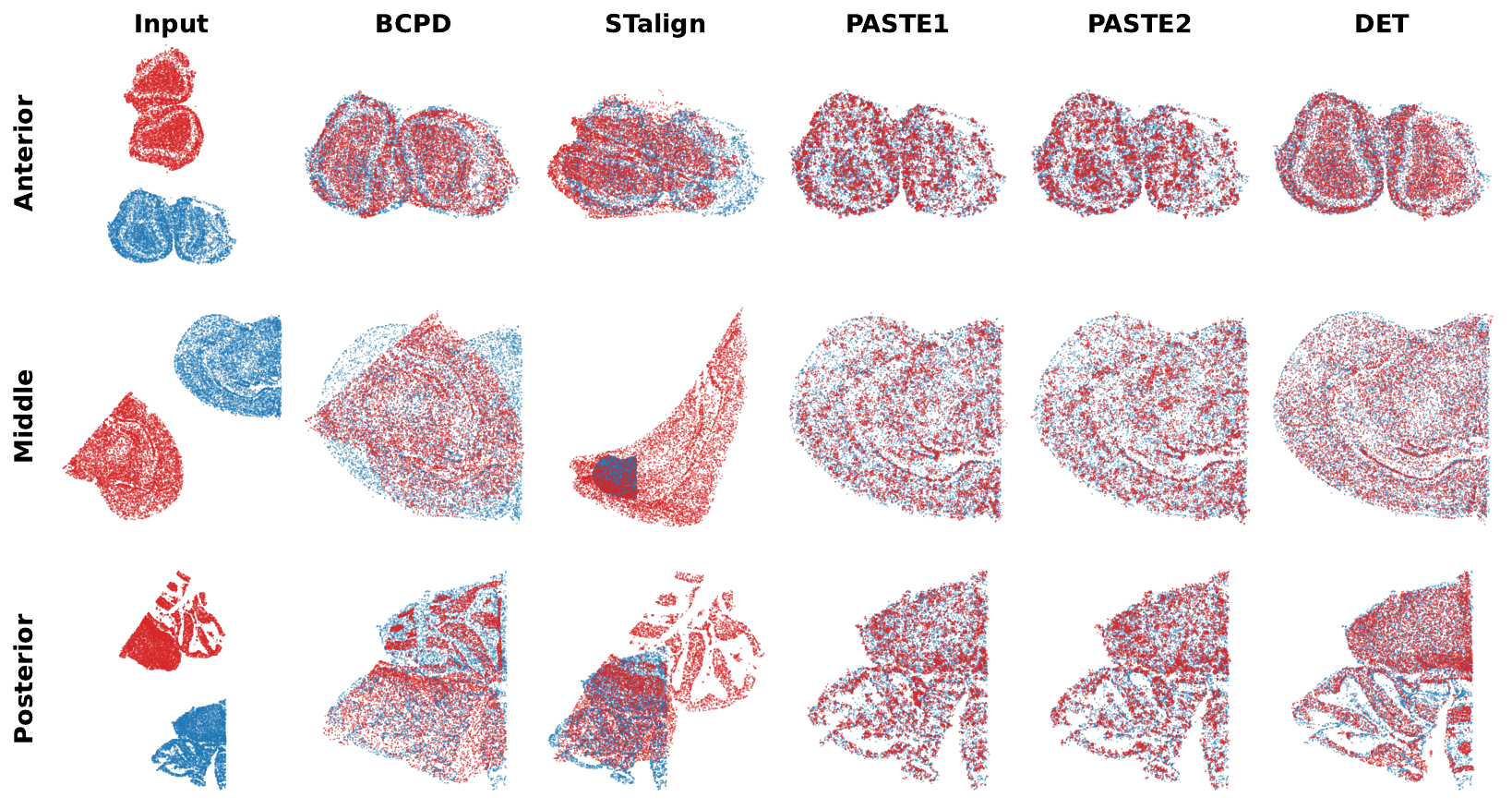}
  \caption{\textbf{DET versus representative baselines on MERFISH slices under severe global perturbation.}
  Each row shows a representative direct registration case from one of three anatomical regions:
  anterior, middle, and posterior. The first column shows the perturbed source slice
  and fixed target slice before registration.  The remaining columns show the registered
  source coordinates overlaid on the target after applying BCPD, STalign,
  the PASTE1-DS-1NN and PASTE2-DS-1NN pipelines (labeled PASTE1 and PASTE2 for
  compactness), and DET. The two PASTE pipelines optimize on
  2000 uniformly sampled landmarks and propagate barycentric landmark displacements
  to all source cells by one-nearest-neighbor interpolation. The overlays are
  illustrative and should not be interpreted as direct measurements of local
  geometric or functional coherence.
  }
  \label{fig:merfish:sota_direct90}
\end{figure*}

\subsubsection{Experimental Setup}

\textbf{Dataset and preprocessing:}
We constructed a 90-case benchmark from three anatomical windows: anterior
slices Zhuang-ABCA-1.004--014, middle slices Zhuang-ABCA-1.083--093, and
posterior slices Zhuang-ABCA-1.136--146. Each window contains 10 adjacent
slice pairs, and each pair was evaluated in three independent random rigid
perturbation trials. For registration, we used cell centroid coordinates and a
shared 50-dimensional global PCA representation of the log-normalized expression
profiles. The PCA basis was fitted once across the benchmark collection and then
applied to every slice, so that the feature axes are shared across all source and
target slices.

\textbf{Robustness protocol:}
To test global registration robustness, we applied a severe synthetic rigid
perturbation to each source slice before registration. The perturbation consisted
of a random rotation sampled over the full range of $0$--$360^\circ$ and a random
translation sampled within the spatial scale of the slice. The target slice was
kept fixed. This protocol forces each method to recover the global pose before
performing local refinement, rather than relying on an approximately initialized overlay. 
Fig.~\ref{fig:merfish:sota_direct90} visualizes representative
source-target pairs used in this direct perturbation protocol and
shows representative qualitative comparisons among the baseline methods.

\textbf{Baselines:}
We compared DET with representative baselines from the method classes most
relevant to our problem setting: 

\begin{itemize}
    \item \textbf{BCPD~\cite{Hirose21,Hirose21a,Hirose23}:}
    A geometry-only baseline. It uses the same class of motion coherence prior
    as DET but does not use the functional likelihood.
    \item \textbf{STalign~\cite{Clifton23STalign}:}
    A diffeomorphic image-registration baseline.
    The sparse point data are rasterized before registration, so this baseline
    represents the grid-based image-registration paradigm.
    \item \textbf{PASTE1-DS-1NN and PASTE2-DS-1NN~\cite{Zeira22,PASTE2GenomeRes23}:}
    Optimal-transport baselines. Because full-resolution runs were
    computationally prohibitive on the 90-case benchmark, we accelerated the methods by
    \begin{enumerate}
    \item  uniformly sampling 2000 landmark cells for each source-target pair,
    \item  running the methods on the landmarks,
    \item  converting the transport plan into landmark displacements by barycentric projection, and
    \item  propagating the displacements to all source cells by one-nearest-neighbor interpolation.
    \end{enumerate}
\end{itemize}

\textbf{Implementation details:}
DET used a two-level hierarchy. The first level performed function-aware
similarity registration for global pose recovery. The second level performed
nonrigid refinement after global alignment. We used the parameters 
shown in Table \ref{tab:param:set}. For PASTE1-DS-1NN/PASTE2-DS-1NN, we used the following parameters:
\begin{itemize}
\item \texttt{n\_landmarks=2000}
\item \texttt{numItermax=10000}
\item \texttt{overlap\_fraction=0.90}  (PASTE2 only)
\end{itemize}
These parameters were selected from a small validation subset because they
allowed all runs to complete and provided adequate validation performance. Optimization
used 2000 landmarks, but all reported metrics were computed on the full cell sets
after displacement interpolation. The results therefore characterize the complete
accelerated landmark-plus-interpolation pipelines evaluated here and should not be
interpreted as full-resolution evaluations of the native PASTE or PASTE2 algorithms.

\subsubsection{Evaluation Metrics}

We evaluated each registered source slice using metrics that measure spatial,
structural, functional, and semantic accuracy.

\begin{enumerate}
    \item \textbf{Jaccard overlap:} We rasterized the registered source and
    target point sets into a shared $50\times50$ occupied-bin grid and computed
    their Jaccard overlap. This metric measures coarse spatial overlap and is
    sensitive to global registration failure.
    \item \textbf{Topology score:} For each source cell, we compared its
    $k=10$ nearest neighbors before and after registration. This metric measures
    preservation of local tissue neighborhoods.
    \item \textbf{Smoothed PCC:} We computed expression agreement after spatial
    smoothing with $k=15$ nearest neighbors and averaged the Pearson correlation
    across registered source cells.
    \item \textbf{Label-transfer ARI:} For each target cell, we transferred the
    label of its nearest registered source cell and computed the adjusted Rand index
    between the transferred labels and the target labels.
    We report \texttt{subclass\_label} as the semantic label.
    \item \textbf{Runtime and failures:} We recorded wall-clock runtime per
    registration and counted a run as failed if it did not produce valid
    coordinates or terminated with an error.
\end{enumerate}

The formal definitions of the metrics can be found in the supplementary material.

\subsubsection{Results and Discussion}

The quantitative results are summarized in Table~\ref{tab:merfish_results}.
All methods completed all 90 registrations. DET achieved the highest
Jaccard overlap, topology score, and smoothed PCC. PASTE2-DS-1NN achieved the
highest subclass-label ARI, indicating strong semantic label-transfer behavior,
but its topology score was substantially lower than that of DET. 
Thus, the main comparison shows a trade-off: PASTE2-DS-1NN is a strong optimal-transport
semantic baseline, whereas DET provides the strongest combination of spatial overlap,
smoothed-expression agreement, and local-neighborhood preservation under the severe
global-perturbation protocol.

\begin{table*}[t]
\begin{center}
\caption{Comparison on the 90-case severe-perturbation MERFISH subclass-label benchmark.}
\label{tab:merfish_results}
\small
\begin{tabular}{lcccccc}
\hline
Method    & $N_{\mathrm{ok}}/N_\mathrm{run}$ &  Jac. $\uparrow$ & Top. $\uparrow$ & Sm. PCC $\uparrow$ & Lt. ARI $\uparrow$ & Time (s) $\downarrow$ \\ \hline
BCPD      & 90/90               &  0.612 $\pm$ 0.288 & 0.897 $\pm$ 0.103 & 0.366 $\pm$ 0.285 & 0.073 $\pm$ 0.085 & \textbf{2.98 $\pm$ 0.61} \\
STalign   & 90/90               &  0.597 $\pm$ 0.247 & 0.782 $\pm$ 0.117 & 0.326 $\pm$ 0.261 & 0.063 $\pm$ 0.086 & 142.00 $\pm$ 5.90 \\
PASTE1-DS-1NN    & 90/90               &  0.865 $\pm$ 0.036 & 0.381 $\pm$ 0.055 & 0.705 $\pm$ 0.121 & 0.181 $\pm$ 0.108 & 3.82 $\pm$ 0.59 \\
PASTE2-DS-1NN    & 90/90               &  0.857 $\pm$ 0.038 & 0.394 $\pm$ 0.052 & 0.719 $\pm$ 0.119 & \textbf{0.194 $\pm$ 0.111} & 9.12 $\pm$ 2.98 \\
DET       & 90/90               &  \textbf{0.881 $\pm$ 0.075} & \textbf{0.928 $\pm$ 0.047} & \textbf{0.736 $\pm$ 0.087} & 0.185 $\pm$ 0.088 & 8.58 $\pm$ 4.39 \\
\hline
\end{tabular}
\end{center} \vspace{1mm}
{\em Values are mean $\pm$ standard deviation over completed
runs. PASTE1-DS-1NN and PASTE2-DS-1NN optimize on 2000 landmarks and
propagate barycentric landmark displacements to all source cells by 1NN
interpolation; metrics are then computed on the full cell sets. Boldface
indicates the top-ranked method for each metric among the listed methods.}
\end{table*}

Fig.~\ref{fig:merfish:sota_direct90} provides representative spatial overlays
from the anterior, middle, and posterior windows and illustrates gross pose
recovery and tissue overlap. Because an overlay alone does not quantify local-neighborhood
preservation or functional agreement, conclusions regarding topology
and functional coherence are based on the quantitative metrics in
Table~\ref{tab:merfish_results}.

\textbf{Robustness to global perturbation:}
BCPD and STalign showed lower overlap and larger variability than DET under
the severe perturbation protocol. BCPD is a geometry-only method and can be
trapped by ambiguous tissue shapes when the initial rotation and translation are large. 
STalign estimates a smooth diffeomorphic deformation after rasterization, but the present
protocol deliberately requires recovery from severe random global pose perturbations.
No separate global initialization was provided. Therefore, the STalign result should
be interpreted as direct recovery by a rasterization-based image-registration pipeline
from a severe random pose, rather than as a general statement about favorably initialized
STalign workflows. In contrast,
DET uses the high-dimensional functional signal during the global stage and achieved
the strongest spatial overlap on the 90-case benchmark.

\textbf{Semantic accuracy and structural preservation:}
PASTE2-DS-1NN achieved the highest subclass-label ARI, and both accelerated PASTE variants
obtained high smoothed PCC values. This indicates that optimal-transport matching is
competitive for semantic label transfer. However, their topology scores were much
lower than those of DET. This should not be interpreted simply as an execution
failure. The observed topology scores reflect the complete evaluated pipeline,
including barycentric landmark displacement and piecewise-constant 1NN interpolation;
they do not isolate the topology of the native optimal-transport solution. These
outputs can align cells semantically while substantially changing local source
neighborhoods. DET instead
regularizes a coherent spatial deformation, yielding substantially higher
topology while maintaining competitive functional and semantic accuracy.

\textbf{Overall interpretation:}
The MERFISH comparison does not show that one method dominates all
metrics. Instead, it clarifies the trade-offs among the evaluated pipelines.
PASTE2-DS-1NN
provides the strongest subclass-level semantic label transfer in this benchmark,
whereas DET provides the strongest combination of spatial overlap,
smoothed-expression agreement, and local-neighborhood preservation. This balance
is the desired behavior for grid-free registration of high-dimensional functions
when both biological correspondence and tissue geometry must be preserved.

\subsubsection{Component Ablations}
We also performed component ablations on a 27-case MERFISH subset; full
numerical results are reported in the supplementary material. These ablations
showed that dimensionality balancing in $\zeta=\eta (D/D')$ is essential, since
the unbalanced diagnostic failed in all 27 cases. 
Functional annealing gave small but consistent gains over a fixed functional weight.
Isotropic and diagonal covariance performed similarly on PCA50, while VGIS and
adaptive outlier modeling had modest effects in this dense MERFISH subset.
Pure geodesic coherence
reduced the topology score in this 2D slice-registration benchmark, so we used Euclidean
coherence for the MERFISH experiments.

\subsubsection{Sensitivity Analyses}
The supplementary material further reports one-factor-at-a-time sensitivity analyses
for the feature representation, functional confidence $\eta$, nonrigid parameters
$\lambda$ and $\beta$, outlier probability $\omega$, landmark budget $M'=N'$,
adaptive-outlier threshold $D_0$, and the smoothing-neighborhood size used for Smoothed PCC.
These studies show that DET is stable over broad practical parameter ranges:
moderate PCA representations give consistent performance, $\lambda$ and $\beta$ mainly
control the flexibility-topology trade-off, low-to-moderate $\omega$ values remain
reliable, and the main conclusions are insensitive to the tested $D_0$ and 
Smoothed PCC neighborhood size.

\subsubsection{Scalability to Large Datasets}
\label{sec:merfish_scalability}

\begin{figure}[t]
\centering
\includegraphics[width=\columnwidth]{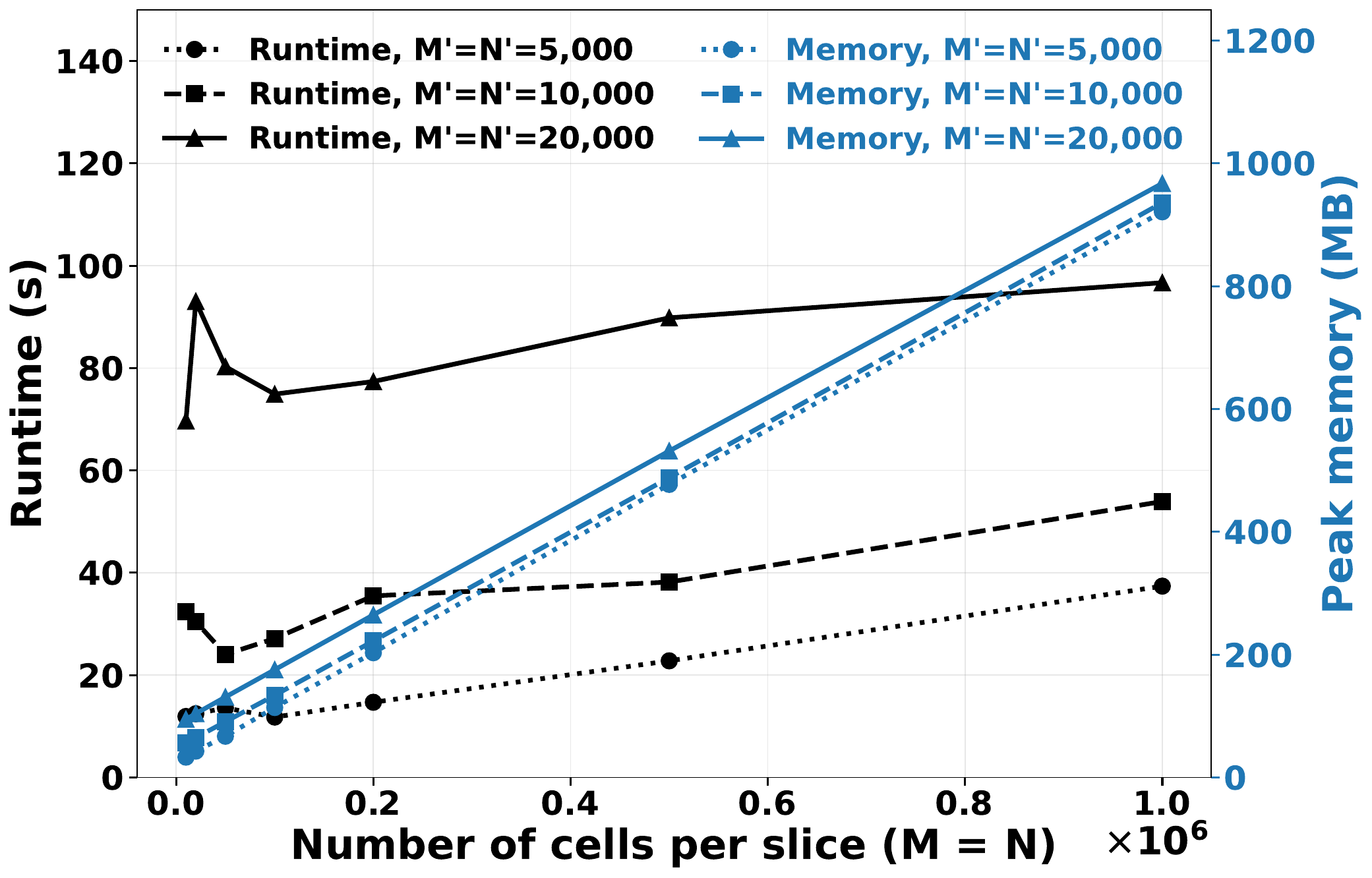}
\caption{Scalability of DET in a synthetic MERFISH cell-count stress test.  Synthetic source and
target slices were generated by resampling the 9-case MERFISH subset to the specified number of
cells.  The L1 similarity stage used $M'=N'=2000$, and the L2 nonrigid stage used
$M'=N'\in\{5000,10000,20000\}$.  Black curves show mean runtime, and blue curves show mean
peak resident memory over the nine base cases.}
\label{fig:det_scalability_runtime_memory}
\end{figure}

To evaluate computational scalability, we performed a synthetic cell-count stress test based on
the 9-case MERFISH subset used in the sensitivity analyses.  For each base source-target pair,
we generated synthetic source and target slices by bootstrap-resampling cells to
$M=N\in\{1,2,5,10,20,50,100\}\times 10^4$, copying the shared PCA50 expression features,
and adding a small amount of coordinate jitter to avoid exact duplicate locations.  This experiment is
intended to measure computational dependence on the full cell count, rather than biological
registration accuracy on new tissue samples.

\textbf{Registration settings:}
DET used the same two-level hierarchy as in Table~\ref{tab:param:set}.  The L1 similarity stage
used a fixed landmark budget of $M'=N'=2000$, because a moderate landmark count was sufficient
for global pose recovery.  The L2 nonrigid stage used $M'=N'\in\{5000,10000,20000\}$ to test
the effect of the refinement-stage landmark count.  We recorded the wall-clock runtime of the
complete hierarchical registration, including interpolation to all source cells.  Because the
underlying C binary does not report memory internally, peak memory was measured externally as
the maximum resident set size of the process during execution.

\textbf{Results:}
Fig.~\ref{fig:det_scalability_runtime_memory} summarizes the results.  With a fixed L2 landmark
budget, runtime increased slowly with the number of cells per slice.  At $M=N=10^6$, the mean
runtimes were 37.4, 53.9, and 96.7 seconds for $M'=N'=5000$, 10000, and 20000, respectively.
Peak memory increased almost linearly with the full cell count and remained below 1~GB on
average in all tested settings; at $M=N=10^6$, the corresponding mean peak resident-memory values
were 921.1, 936.0, and 967.4~MB.  These results are consistent with the intended acceleration
design: the iterative optimization is governed mainly by the downsampled landmark set, whereas
the costs of full-resolution interpolation and storage grow approximately linearly with $M+N$.

The experiment therefore supports the practical scalability of DET for atlas-scale cell sets.  The
landmark count provides an explicit trade-off between computational cost and refinement detail,
while the dependence on the full cell count remains controlled by the downsampling and
interpolation strategy.

\subsection{Atlas-Scale Cross-Stage Consistency (Stereo-seq)}
\label{sec:exp:msta}

\begin{figure*}[t]
  \centering
  \includegraphics[width=\linewidth]{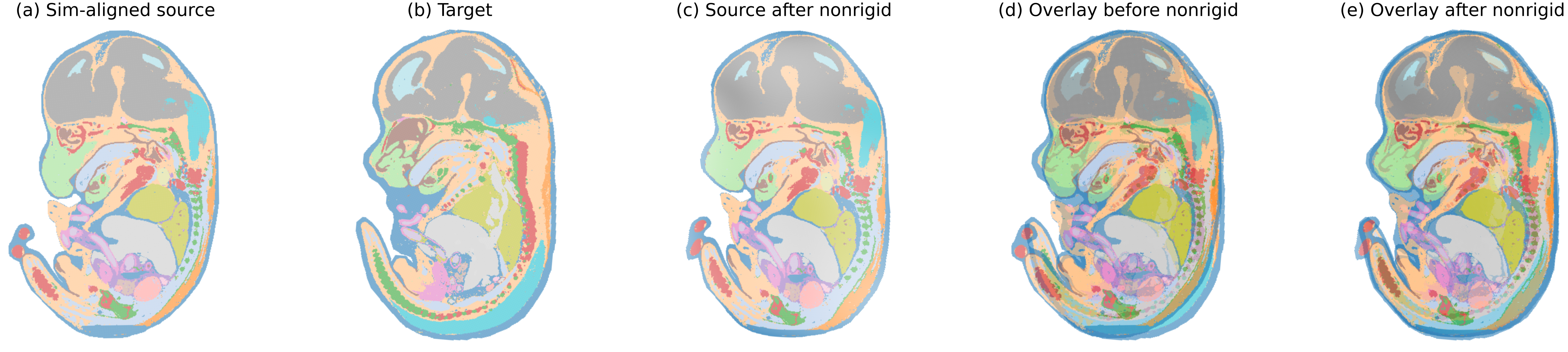}
  \caption{\textbf{Cross-stage MOSTA registration from E14.5 to E15.5.}
  (a) Similarity-aligned E14.5 source, (b) E15.5 target, and (c) source after
  L2 nonrigid refinement. Points are colored by anatomical-domain annotations.
  (d) Overlay before nonrigid refinement and (e) overlay after nonrigid refinement.
  The overlay panels illustrate the spatial effect of L2 refinement after global pose recovery.
  }
  \label{fig:mosta_reg}
\end{figure*}

To demonstrate the scalability of DET to atlas-level data and its ability to handle
complex developmental deformation, we applied the algorithm to mouse embryo
Stereo-seq data (MOSTA)~\cite{Chen22}. Unlike the MERFISH benchmark in Section~\ref{sec:exp:merf},
which aligns nearby brain sections with comparable cell-type annotations,
this experiment considers a more challenging
\textit{cross-stage} setting: registering an E14.5 embryo to an E15.5 embryo.
Because cell populations and expression profiles change during development and
no curated cross-stage correspondence ground truth or directly comparable baseline
result is available, we use the MOSTA experiment as an atlas-scale feasibility and within-pipeline
consistency study rather than as an accuracy benchmark.

\textbf{Challenge:}
The primary difficulty stems from substantial biological growth and morphological
change over the 24-hour developmental interval. Cross-stage alignment requires the
transformation to handle global pose differences and nonrigid growth of developing
organs. We therefore evaluate the result using anatomical-domain and boundary
metrics that match this setting, in addition to global overlap, topology, and local
expression consistency.

\textbf{Dataset and preprocessing:}
We used the E14.5 sagittal section as the source point set and the E15.5 section
as the target. The source and target comprised 102,519 and 113,350 points, respectively. To ensure
that the functional coordinates were comparable across developmental stages, PCA
was not fitted separately to the two embryos. Instead, we took the intersection
of the two gene sets, applied the same total-count normalization and log transformation,
concatenated the E14.5 and E15.5 expression matrices, and fitted a
single shared PCA basis.
The projected scores on the first 50 shared principal components were then split back
into source and target feature matrices and used as $F_y$ and $F_x$.
For the final MOSTA setting, the L2 stage used VGIS. To directly check the contribution of VGIS,
we also evaluated a matched L2 variant that used uniform downsampling while keeping
the same acceleration settings.

\textbf{Metrics:}
We report diagnostics that capture different aspects of cross-stage
consistency. Occupancy Jaccard measures global tissue overlap, topology measures
whether local source neighborhoods are preserved, and Smoothed PCC measures local
expression agreement. Common-domain accuracy, macro-F1, and ARI summarize coarse
anatomical-domain agreement after nearest-neighbor label transfer, with the evaluation restricted
to domains shared by E14.5 and E15.5. Boundary F1, boundary Jaccard, and boundary
Chamfer summarize interface consistency, while domain Chamfer and domain-centroid
error summarize domain placement. These quantities compare the initial, L1,
L2-uniform, and L2-VGIS configurations within the DET pipeline; without
correspondence ground truth, they do not establish exact developmental accuracy or
superiority over other algorithms. Formal definitions are given in the supplementary material.

\begin{table}[t]
\centering
\caption{MOSTA E14.5-to-E15.5 within-pipeline consistency diagnostics. These
values compare DET configurations and are not ground-truth accuracy measures.}
\label{tab:mosta_quant}
\tiny
\setlength{\tabcolsep}{2.0pt}
\renewcommand{\arraystretch}{0.88}
\resizebox{\columnwidth}{!}{%
\begin{tabular}{lcccc}
\hline
Metric & Initial & L1 sim. & L2 uniform & L2 + VGIS \\
\hline
Occupancy Jaccard $\uparrow$   & 0.000 & 0.901   & \textbf{0.971} & 0.964 \\
Topology $\uparrow$            & \textbf{1.000}  & 0.902 & 0.883 & 0.885 \\
Smoothed PCC $\uparrow$        & -0.381  & 0.082 & 0.124 & \textbf{0.126} \\
Common Acc. $\uparrow$         & 0.058   & 0.413 & 0.468 & \textbf{0.471} \\
Common macro-F1 $\uparrow$     & 0.005   & 0.333 & \textbf{0.351} & 0.347 \\
Common ARI $\uparrow$          & 0.000   & 0.289 & 0.345 & \textbf{0.350} \\
Boundary F1 $\uparrow$         & --      & 0.399 & 0.442 & \textbf{0.446} \\
Boundary Chamfer $\downarrow$  & 608.959 & 2.824 & 2.190 & \textbf{2.122} \\
Boundary Jaccard $\uparrow$    & 0.000 & 0.439 & \textbf{0.486} & 0.486 \\
Domain Chamfer $\downarrow$    & 650.099 & 10.872 & 10.076 & \textbf{9.954} \\
Domain centroid error $\downarrow$   & 825.844 & 43.583 & \textbf{41.343} & 41.412 \\
\hline
\end{tabular}%
}
\end{table}

\begin{figure}[t]
\centering
\includegraphics[width=\columnwidth]{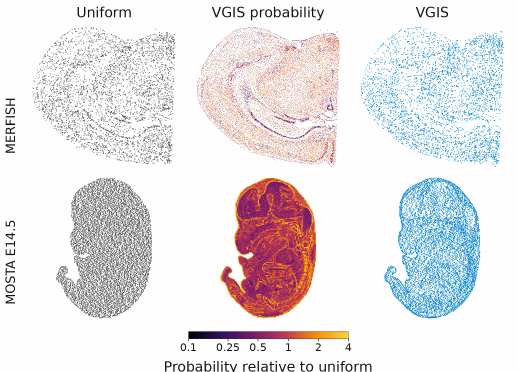}
\caption{\textbf{Dataset-dependent behavior of VGIS.}
Each row shows uniform landmarks, the VGIS sampling-probability map, and VGIS
landmarks for representative MERFISH and MOSTA data. Colors report pointwise
probability relative to uniform sampling, $M p(z_i)$, on a common logarithmic scale clipped to $[0.1,4]$.
MERFISH importance is comparatively diffuse, whereas MOSTA contains 
more strongly localized regions of functional variability and outer-contour emphasis.
}
\label{fig:vgis_sampling_comparison}
\end{figure}

\textbf{Results:}
Table~\ref{tab:mosta_quant} shows the within-pipeline consistency diagnostics, and
Fig.~\ref{fig:mosta_reg} visualizes the registration process, with points
colored by anatomical-domain annotations.
The L1 similarity stage recovered a compatible global pose, increasing occupancy
Jaccard from 0.000 to 0.901. The L1 topology score of 0.902, despite L1 being a
global similarity transform, results from $k$-nearest-neighbor tie-breaking on the
regularly sampled Stereo-seq lattice after coordinate rounding, rather than from
local deformation. L2 nonrigid refinement further improved several
coarse consistency measures, with the
VGIS setting reaching a common-domain ARI of 0.350 and reducing 
boundary Chamfer from 2.824 after L1 to 2.122.

Fig.~\ref{fig:vgis_sampling_comparison} visualizes the landmark distributions used in L2
alongside a representative MERFISH comparison.
Uniform downsampling covers the embryo nearly evenly, whereas VGIS places
more landmarks along the outer contour and in localized functional-variability regions.
The probability map should not be interpreted as an anatomical-boundary annotation.
The sampling difference is consistent with the trend in Table~\ref{tab:mosta_quant}:
compared with uniform downsampling, VGIS improves common-domain ARI, boundary F1,
boundary Chamfer, domain Chamfer, topology, and Smoothed PCC. Uniform downsampling gives slightly higher
occupancy Jaccard and common-domain macro-F1, and slightly lower domain-centroid error.
Thus, VGIS does not simply maximize global coverage; rather, it improves several
boundary- and domain-consistency measures relevant to cross-stage anatomical alignment.

The further topology decrease from L1 to L2 is expected: a global similarity
transform preserves local
neighborhoods, whereas nonrigid refinement must deform the E14.5
embryo to match the more developed E15.5 target. Because no curated cross-stage
correspondence ground truth or directly comparable baseline result is available,
these changes should be interpreted as evidence of atlas-scale feasibility and
increased within-pipeline consistency. They do not establish exact developmental
correspondence or general superiority over alternative cross-stage methods.

\section{Discussion}

This section clarifies the methodological differences from BCPD
and then discusses trade-offs, implications, and limitations revealed
by the complementary MERFISH and MOSTA experiments.

\subsection{Methodological Divergence from BCPD}
\label{sec:det_bcpd}

The proposed DET framework represents a shift from geometric point-set
registration to function registration on irregular domains. It inherits the core
BCPD machinery, i.e., variational inference, latent correspondences, the motion
coherence prior, and acceleration schemes, from BCPD~\cite{Hirose21,Hirose21a,Hirose23}.
Table~\ref{tab:bcpd_vs_det} summarizes the core conceptual and algorithmic
differences. The transition from pure spatial coordinates to functions introduces
optimization challenges, which DET addresses through several methodological components.

\begin{table}[h]
  \caption{Relationship between BCPD and DET.}
  \label{tab:bcpd_vs_det}
  \begin{center}
  \footnotesize
  \begin{tabular}{@{}p{0.25\columnwidth}p{0.30\columnwidth}p{0.35\columnwidth}@{}} \hline
                             & \textbf{BCPD} \cite{Hirose21,Hirose21a,Hirose23}  & \textbf{DET (Proposed)} \\ \hline
    {\bf Problem}            & Point Set Alignment    & {\bf Function Alignment   }         \\
    {\bf Input}              & Point sets             & {\bf Discretized Functions}         \\
    {\bf Likelihood}         & Spatial                & {\bf Spatial-Functional   }         \\
    Inference                & Variational Inference  & Variational Inference               \\
    Motion Prior             & Gaussian Process       & Gaussian Process                    \\
    {\bf Acceleration}       &                        &                                     \\
    ~~  Fast $G/P$           & Nystr\"om + $k$-D tree & Nystr\"om + $k$-D tree              \\
    ~~  {\bf Downsampling}   & Uniform                & {\bf VGIS/Uniform}$^\dagger$        \\
    ~~  Interpolation        & GP Regression          & GP Regression                       \\
    {\bf $\zeta$-Balancing}  & ~~~~~~~     --         & ~~~~~~~     \checkmark              \\
    {\bf Functional Cov.}    & ~~~~~~~     --         & {\bf Diagonal/Isotropic}            \\
    {\bf Outlier Dist.}      & Uniform                & {\bf Adaptive/Uniform}$^\dagger$    \\
    {\bf Annealing}          & Coherence              & {\bf Coherence + Weight $\eta$}     \\
    \hline
  \end{tabular} 
  \end{center}
  {\em Bold entries indicate differences between DET and BCPD; VGIS and the adaptive
  outlier model (with $\dagger$ symbols) are optional choices.}
\end{table}

The following components address difficulties that arise when moving from
geometric point-set registration to function registration on irregular domains.
The spatial-functional likelihood, likelihood balancing, and functional annealing are central
to the evaluated DET hierarchy. 
DET uses diagonal covariance as its default tractable parameterization,
with isotropic covariance as a restricted alternative. Adaptive outliers,
feature-sensitive sampling, and alternative coherence matrices are optional
choices whose benefits depend on the data.

{\bf 1) Spatial-Functional Likelihood.}
BCPD maximizes the likelihood of spatial proximity alone.
In contrast, DET combines spatial proximity and functional similarity through
the local model factor in Eq.~(\ref{eq:joint_likelihood}), which is used in the
generalized mixture construction of Eq.~(\ref{eq:generative_model}). 
This is not merely a matter of appending a feature column; it couples spatial alignment with functional identity.
This allows DET to leverage non-spatial cues to resolve geometric ambiguities caused by rotational symmetry in tissue geometry.

{\bf 2) Likelihood Balancing.}
Simply appending functional dimensions in a standard Bayesian framework is
problematic due to the curse of dimensionality. High-dimensional signals, such as gene
expression profiles ($D' \approx 1{,}000$), can dominate the spatial term. DET addresses 
this through the likelihood balancing parameter $\zeta = \eta(D/D')$
(Eq.~\ref{eq:zeta_balancing}), which balances the spatial and functional energy scales.

{\bf 3) Tractable Functional Covariance.}
DET uses diagonal $\Pi$ as the default tractable covariance model for high-dimensional
registration, avoiding unstable cross-feature covariance estimation while retaining
feature-wise adaptation. Isotropic covariance is a restricted special case with shared variance.
On PCA50 MERFISH, diagonal covariance gave no accuracy gain, suggesting limited benefit
from feature-wise variance adaptation in this compact PCA representation,
although our normalization preserves unequal component variances.
We therefore regard it primarily as a stable general parameterization;
its benefit for raw or minimally reduced heterogeneous features remains untested.

{\bf 4) Adaptive Outlier Model.}
DET provides the feature-aware model in Eq.~(\ref{eq:adaptive_outlier}) as an alternative
to BCPD's uniform outlier distribution. 
Its scale is matched to the weighted spatial-functional likelihood.
The adaptive and uniform formulations perform similarly on the
PCA50 MERFISH data, so we regard this as an optional dimension-aware
modeling choice rather than an established performance driver.

{\bf 5) Feature-Sensitive Sampling.}
Whereas BCPD typically employs uniform downsampling, such an approach can spend
most landmarks on homogeneous interior area when informative interfaces or rare
high-variability regions occupy only a small fraction of a large point cloud. DET
provides VGIS, which computes a sampling probability
(Eq.~\ref{eq:vgis_probability}) based on local functional variability and geometric
discontinuity. When variability is broadly distributed, VGIS can approach uniform
sampling and need not improve registration.

{\bf 6) Functional Annealing.}
Standard hierarchical BCPD anneals only the motion coherence parameters
($\lambda$, $\beta$). DET introduces a dual-annealing strategy termed functional
annealing (Eq.~\ref{eq:functional_annealing}). By initializing with high functional
confidence $\eta$ to help resolve global ambiguities, e.g., rotation and translation, and
progressively shifting priority to spatial fidelity, 
DET helps avoid overfitting points to noisy functional data---a failure mode
particularly relevant to functional registration.

These methodological divergences help explain DET's observed behavior in complex
biological settings, such as slice-to-slice MERFISH alignment.

\subsection{Complementary Roles of MERFISH and MOSTA}
\label{sec:discussion_merfish_mosta_roles}

The MERFISH and MOSTA experiments evaluate different aspects of DET. The MERFISH benchmark
is a controlled adjacent-slice problem. The source and target have comparable anatomy,
expression profiles, and cell-type annotations. It therefore tests whether functional
information helps recover severe pose perturbations while preserving tissue geometry.
The MOSTA experiment is instead a cross-developmental whole-embryo alignment from E14.5 to
E15.5. It tests whether the same grid-free registration framework remains practical and
anatomically plausible at atlas scale, even when expression and tissue composition can change
with development.

\subsubsection{Trade-offs Observed in the Experiments}
\label{sec:discussion_what_results_show}

The MERFISH results show the main benefit of adding functional information to point set registration.
Geometry-only registration can preserve shape, but it has difficulty resolving ambiguous tissue geometry
under large rotations and translations. Image-based registration is a natural alternative, but it
requires rasterization of sparse cell coordinates. Our STalign experiment specifically tests direct
recovery from a severe random pose without a separate global initialization and does not characterize
favorably initialized STalign workflows. Optimal-transport methods provide
strong semantic matching, and the evaluated PASTE2-DS-1NN pipeline is competitive
in label-transfer ARI. Its observed local-neighborhood changes reflect the complete
barycentric-landmark plus 1NN-interpolation pipeline and should not be attributed
solely to the native PASTE2 algorithm. DET occupies a different part
of this trade-off; it uses expression to guide correspondence while estimating a coherent deformation
of the spatial domain.

The MERFISH ablations suggest that the functional likelihood is the central ingredient, whereas
some engineering components have more dataset-dependent effects. Likelihood balancing is
essential, not merely helpful; when the $D/D'$ factor in $\zeta=\eta (D/D')$ was removed in the
unbalanced diagnostic, all 27 MERFISH ablation cases failed to produce valid registered coordinates.
Functional annealing is also useful; strong functional guidance helps recover the global pose,
whereas a milder functional weight during nonrigid refinement reduces the risk of following
noisy expression variation too aggressively.
Thus, the qualitative failure mode is asymmetric. Too small a value of $\eta$ makes DET approach
geometry-only BCPD, whereas too large a value of $\eta$ can make the functional term dominate the
nonrigid stage, leading to overfitting to noisy expression variation or numerical instability.

VGIS shows a different, dataset-dependent pattern. In the dense adjacent MERFISH slices,
functional variability is broadly distributed, so uniform landmarks already cover the relevant
regions, and there is relatively little homogeneous interior area for VGIS to deprioritize. 
In MOSTA, uniform downsampling covers the whole embryo well and can give slightly higher global occupancy overlap,
whereas VGIS concentrates landmarks near the outer contour and localized functional-variability
regions. In the selected setting, VGIS improves several boundary- and domain-oriented diagnostics.
We therefore interpret VGIS not as universally superior to uniform sampling, but as potentially
useful when large irregular point clouds contain extensive homogeneous interiors and comparatively
sparse anatomical interfaces that may be underrepresented by a limited uniform landmark budget.
Fig.~\ref{fig:vgis_sampling_comparison} illustrates this contrast: MERFISH has a relatively
diffuse importance pattern, whereas MOSTA has a more strongly localized probability distribution. These probability
maps are not anatomical-boundary annotations, however, and localized feature variation may reflect
either biological or technical signals; the maps alone therefore do not establish that every
high-probability region is an anatomical interface or that VGIS improves registration accuracy.

\subsubsection{Implications and Limitations}
\label{sec:discussion_implications_limitations}

The experiments suggest that DET is most useful when geometry alone is ambiguous and
rasterized image registration would discard important point-level information. The method is
not meant to replace all spatial-transcriptomics alignment tools. Optimal-transport methods may
be preferable when semantic matching is the primary objective and local geometric coherence is
less important. DET is preferable when the output should be a smooth deformation of the source
domain and when high-dimensional features are needed to disambiguate correspondence without
voxelizing the data.

Several limitations remain. 
First, the current experiments use datasets whose source and target features can be represented
in a shared coordinate system, such as a shared gene panel or a jointly fitted PCA
basis. Non-shared gene panels require a common embedding or feature-matching method
before DET can be applied.
Second, MOSTA lacks standardized cross-stage ground truth or a directly comparable
baseline; its metrics should be interpreted
as anatomical consistency checks rather than as proof of exact developmental correspondence.
Third, the independent functional normalization used in our experiments focuses the registration
on relative spatial expression patterns. This choice is useful for mitigating batch and 
intensity-scale differences, but it may remove global expression differences that are biologically
meaningful in cross-condition studies, such as healthy versus diseased tissue.
In such settings, the normalization strategy should be chosen according to the
biological question, for example by using joint, reference-based, or calibrated
normalization.

For spatial-transcriptomics registration, PCA10--PCA50 is our recommended default
range. PCA10 prioritizes speed and denoising, whereas PCA50 retains a broader
expression representation at moderate cost; PCA100 produced no clear improvement
in this benchmark. Full-gene input is not automatically more informative because
noisy or redundant dimensions can dilute correspondence evidence. It should be
reserved for applications in which gene-specific dimensions are scientifically
important and the additional computational cost and noise sensitivity are acceptable.
DET uses diagonal covariance by default for stable high-dimensional inference;
isotropic covariance performed similarly on PCA50, and any benefit for raw or
minimally reduced features remains unverified.

Future work should address registration with features in non-shared coordinate
systems, develop normalization strategies that preserve biologically meaningful
cross-condition differences, improve robustness for extremely sparse
high-dimensional data, explore parallel implementations for ultra-large atlases,
and establish benchmarks with curated landmarks or anatomical correspondences
for cross-stage spatial-omics registration.

\section{Conclusion}

Nonrigid registration is a fundamental problem in pattern analysis, essential
for organizing and interpreting complex data across domains. Historically, the
field has been bifurcated into point set registration, which handles sparse
geometry but ignores functional signals, and image registration, which leverages
intensity fields but requires regular grids. This dichotomy has become a
limitation for emerging scientific data, particularly in spatial transcriptomics,
where high-dimensional vector-valued functions are defined on sparse, irregular
manifolds.

In this study, we proposed DET to bridge this gap. By formulating the problem
as function registration within a generalized Bayesian framework, we derived an algorithm
that registers high-dimensional signals directly on their native domains. The
grid-free formulation avoids the binning process required by standard image
registration and therefore helps preserve point-level scientific information.

Our experiments demonstrated the utility of DET in spatial-transcriptomics
settings. On MERFISH, the method provided a strong balance of overlap, expression
agreement, and topology under severe perturbations without prior initialization. 
The MOSTA study demonstrated atlas-scale cross-stage feasibility and showed that
nonrigid refinement improved several within-pipeline consistency measures.
Together, the results indicate that combining
spatial and functional information can support registration while preserving
coherent domain deformation.

We believe that DET provides a useful tool for scientific registration problems
in low-data regimes. By enabling resolution-preserving registration without
manual annotation or pre-training, DET opens new avenues for analyzing the complex,
multimodal data structures that define modern computational biology and pattern recognition.

\bibliographystyle{IEEEtran}

\bibliography{IEEEabrv,fr-ieee-new}

\begin{IEEEbiography}[{\includegraphics[width=1in,height=1.25in,clip,keepaspectratio]{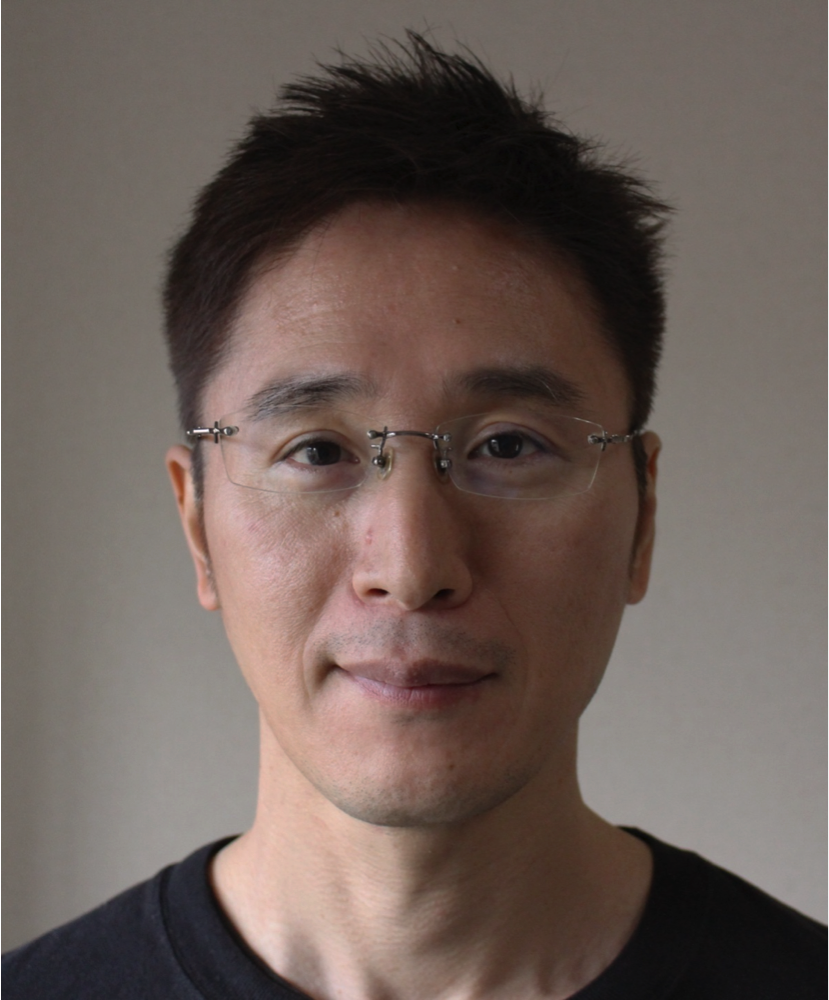}}]{Osamu Hirose}
is currently an Associate Professor with the Institute of Science and Engineering,
Kanazawa University, where he leads the statistical machine learning laboratory. He is also a FOREST Researcher
with the Japan Science and Technology Agency (JST). He received the Ph.D. degree in information science
and technology from the University of Tokyo, Tokyo, Japan, in 2008. 
His research interests include probabilistic registration, statistical shape analysis, computational geometry,
machine learning, statistics, and their applications to biological and scientific data.
He is a member of the IEEE.
\end{IEEEbiography}

\begin{IEEEbiography}[{\includegraphics[width=1in,height=1.25in,clip,keepaspectratio]{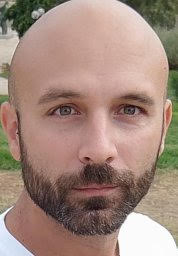}}]{Emanuele Rodol\`a}
is a Full Professor of Computer Science at Sapienza University of Rome,
where he leads the GLADIA group of Geometry, Learning \& Applied AI. His research focuses on
Representation Learning, Machine Learning for Audio, LLMs, Geometric Deep Learning, and Computer Vision.
He is an ERC grantee, a Google Research awardee, and a fellow of both ELLIS and the Young Academy of Europe.
Professor Rodol\`a earned his PhD from Università Ca’ Foscari Venezia in 2012. His career
includes international experience as an Alexander von Humboldt Fellow at TU Munich
and a JSPS Research Fellow at The University of Tokyo. With an h-index of 50 and over 13,000 citations,
he has received numerous Best Paper Awards at premier venues. His research has been featured
in major international media, including RAI, Wired, and La Repubblica.
\end{IEEEbiography}


\end{document}